\documentclass{ieeeaccess}

\usepackage{amsmath,amssymb,amsfonts}
\usepackage{algorithmic}
\usepackage{graphicx}
\usepackage{textcomp}
\usepackage{adjustbox}
\usepackage{subcaption}
\usepackage{booktabs}
\usepackage{multirow}
\usepackage{array}
\usepackage{colortbl} 
\usepackage[table,xcdraw]{xcolor} 

\usepackage{siunitx}
\definecolor{lightgray}{gray}{0.9}
\newcommand{\hide}[1]{}

\newcommand{\bdmath}{\begin{dmath}}
\newcommand{\edmath}{\end{dmath}}
\newcommand{\beq}{\begin{equation}}
\newcommand{\eeq}{\end{equation}}
\newcommand{\bdm}{\begin{displaymath}}
\newcommand{\edm}{\end{displaymath}}
\newcommand{\bea}{\begin{eqnarray}}
\newcommand{\eea}{\end{eqnarray}}
\newcommand{\beal}{\beq \begin{array}{ll}}
\newcommand{\eeal}{\end{array} \eeq}
\newcommand{\beas}{\begin{eqnarray*}}
\newcommand{\eeas}{\end{eqnarray*}}
\newcommand{\ba}{\begin{array}}
\newcommand{\ea}{\end{array}}
\newcommand{\eit}{\end{itemize}}
\newcommand{\ben}{\begin{enumerate}}
\newcommand{\een}{\end{enumerate}}

\newcommand{\noise}{\boldsymbol\eta}

\newcommand{\toa}{ToA\xspace}

\usepackage{xspace}
\usepackage{float}
\usepackage{hyperref}

\usepackage[numbers]{natbib}

\newcounter{subsubsubsection}[subsubsection]

\definecolor{maxred}{RGB}{255,0,0}
\definecolor{maxgreen}{RGB}{0,102,0}

\definecolor{maxred}{RGB}{255,0,0}
\definecolor{lightred}{RGB}{255,51,51}
\definecolor{lightgreen}{RGB}{0,153,0}
\definecolor{medgreen}{RGB}{0,128,0}
\definecolor{deepgreen}{RGB}{0,102,0}

\definecolor{posgreen}{RGB}{34,139,34}    
\definecolor{negred}{RGB}{178,34,34}      

\definecolor{posgreen}{RGB}{0,100,0}      
\definecolor{negred}{RGB}{139,0,0}        

\newcommand{\coloredpercent}[1]{%
    \ifdim#1pt>0pt
        \textcolor{posgreen}{#1\%}%
    \else
        \textcolor{negred}{#1\%}%
    \fi
}

\def\BibTeX{{\rm B\kern-.05em{\sc i\kern-.025em b}\kern-.08em
    T\kern-.1667em\lower.7ex\hbox{E}\kern-.125emX}}
\begin{document}
\history{Date of publication xxxx 00, 0000, date of current version xxxx 00, 0000.}
\doi{10.1109/ACCESS.2017.DOI}

\title{Global SLAM Using 5G ToA Integration: Performance Analysis with Unknown Base Stations and Loop Closure Alternatives}
\author{\uppercase{Meisam Kabiri}\authorrefmark{1},
\uppercase{Holger Voos\authorrefmark{1, 2}}
}
\address[1]{Interdisciplinary Center for Security Reliability and Trust (SnT), University of Luxembourg, Luxembourg }
\address[2]{Faculty of Science, Technology, and Medicine (FSTM), Department of Engineering, University of Luxembourg, Luxembourg }
\tfootnote{This research was funded in whole, or in part, by the Luxembourg National Research 
Fund (FNR), 5G-Sky Project, ref. C19/IS/13713801/5G-Sky/Ottersten. For the purpose of open access, and in fulfilment of the obligations arising from the grant agreement, the authors have applied a Creative Commons Attribution 4.0 International (CC BY 4.0) license to any  Author Accepted Manuscript version arising from this submission.
}


\markboth
{Kabiri \headeretal: Global SLAM in Visual-Inertial Systems with 5G ToA Integration}
{Kabiri \headeretal: Global SLAM in Visual-Inertial Systems with 5G ToA Integration}

\corresp{Corresponding author: Meisam Kabiri (e-mail: meisam.kabiri@uni.lu).}

\begin{abstract}

This paper presents a novel approach that integrates 5G Time of Arrival (ToA) measurements into ORB-SLAM3 to enable global localization and enhance mapping capabilities for indoor drone navigation. We extend ORB-SLAM3's optimization pipeline to jointly process ToA data from 5G base stations alongside visual and inertial measurements while estimating system biases. This integration transforms the inherently local SLAM estimates into globally referenced trajectories and effectively resolves scale ambiguity in monocular configurations. Our method is evaluated using both Aerolab indoor datasets with RGB-D cameras and the EuRoC MAV benchmark, complemented by simulated 5G ToA measurements at 28 GHz and 78 GHz frequencies using MATLAB and QuaDRiGa. Extensive experiments across multiple SLAM configurations demonstrate that ToA integration enables consistent global positioning across all modes while maintaining local accuracy. For monocular configurations, ToA integration successfully resolves scale ambiguity and improves consistency. We further investigate scenarios with unknown base station positions and demonstrate that ToA measurements can effectively serve as an alternative to loop closure for drift correction. We also analyze how different geometric arrangements of base stations impact SLAM performance. Comparative analysis with state-of-the-art methods, including UWB-VO, confirms our approach's robustness even with lower measurement frequencies and sequential base station operation. The results validate that 5G ToA integration provides substantial benefits for global SLAM applications, particularly in challenging indoor environments where accurate positioning is critical.

\end{abstract}


\titlepgskip=-15pt

\maketitle

\section{Introduction}
\label{sec:introduction}
\PARstart{S}{imultaneous} Localization and Mapping (SLAM) systems are essential for autonomous robotics, enabling robots to construct maps of their environments while simultaneously determining their positions within those maps \cite{durrant2006simultaneous}. SLAM technologies are extensively utilized in fields such as autonomous navigation \cite{grisetti2010tutorial}, augmented reality \cite{sheng2024review}, and various robotic applications \cite{bresson2017simultaneous}. Most existing methods, primarily operate within a local reference frame, making it challenging to achieve global localization, which is critical for large-scale environments, multi-agent systems, and integration with external infrastructure \cite{mur2017orb}.

Integrating a global reference frame into SLAM systems addresses these challenges by anchoring localization to a consistent and absolute coordinate system. This enhancement improves positional accuracy, enables seamless interaction with external infrastructure, and supports collaborative multi-agent mapping \cite{zou2019collaborative,karrer2018towards}. Additionally, a global reference frame allows SLAM systems to incorporate supplementary data sources, increasing reliability in scenarios where visual and inertial cues may be insufficient \cite{leitinger2017factor}. This integration also makes SLAM systems more scalable and adaptable for diverse real-world applications \cite{karfakis2023nr5g}.

\begin{figure}
    \includegraphics[width=\linewidth]{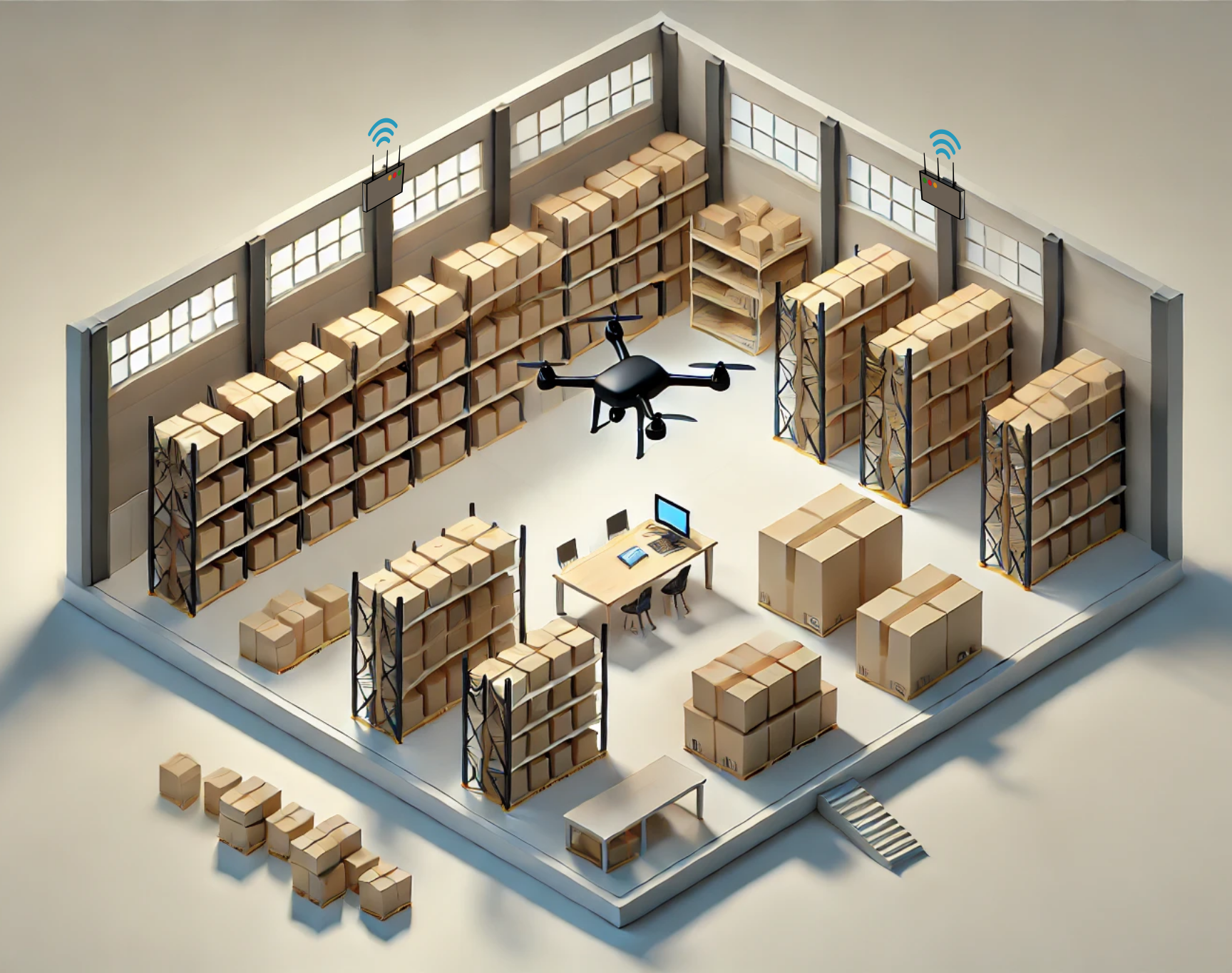}
    \caption[Warehouse navigation with 5G ToA-based localization]{Illustration of an indoor warehouse environment where a drone navigates while two 5G base stations provide \toa measurements. The base stations assist in refining the drone's positional accuracy, supporting real-time localization and mapping (The warehouse figure was generated using OpenAI's ChatGPT).}
    \label{fig:warehouse}
\end{figure}

The advent of 5G technology has introduced advanced positioning capabilities, particularly through \toa measurements \cite{ferre2019positioning, kabiri2022review}. ToA provides highly accurate distance estimates between receivers and known base station positions, offering centimeter-level precision in ideal conditions. By incorporating ToA data, SLAM systems can align their local frames with a global reference frame defined by the fixed positions of 5G base stations. This global alignment enables robust localization and mapping in GPS-denied environments, enhancing applications like inventory management, real-time monitoring, autonomous vehicles, augmented reality, and emergency response. Drones, for instance, can autonomously navigate warehouses, accurately mapping the environment and tracking inventory levels, as illustrated in Fig.~\ref{fig:warehouse}. By leveraging ToA measurements from 5G base stations as global anchors, drones achieve higher accuracy and robustness, optimizing inventory management and real-time monitoring processes.

This paper extends ORB-SLAM3 \cite{campos2021orb} by integrating 5G  \toa measurements into its optimization pipeline to achieve globally consistent SLAM. Specifically, we introduce an SE3 transformation node into the estimation process, deeply integrating it into both the front-end and back-end modules of ORB-SLAM3. This SE3 node estimates the rigid body transformation between the local SLAM reference frame and the global 5G-based reference frame in real-time, aligning the local map with the global coordinate system to ensure precise localization and mapping across different environments. Furthermore, we investigate challenging scenarios with unknown base station positions and limited visibility, demonstrating the system's adaptability to real-world deployment constraints through sequential base station operation.

Additionally, we incorporate bias nodes for each 5G base station to account for clock biases inherent in ToA measurements. These bias nodes are integrated into the optimization graph, allowing the system to jointly estimate the clock biases alongside the transformation between frames. This comprehensive integration of SE3 transformation and bias estimation ensures that the SLAM system maintains high performance and robustness, even in scenarios where traditional visual-inertial SLAM may encounter challenges such as feature-sparse environments or dynamic changes.

We assess the effectiveness of our approach using five indoor datasets captured with RGB-D cameras and Inertial Measurement Units (IMUs) from our Aerolab facility, as well as the standard EuRoC MAV benchmark for comparison with state-of-the-art methods. These datasets are further enhanced with simulated 5G \toa measurements at frequencies of 28 GHz and 78 GHz, generated using MATLAB and QuaDRiGa. The experimental results indicate that our method maintains high local estimation precision, benefiting from the accuracy inherent in  ORB-SLAM3. Moreover, the incorporation of ToA measurements enables global localization and mapping. We also demonstrate that ToA integration can effectively serve as an alternative to loop closure mechanisms for drift correction, maintaining trajectory consistency even in challenging scenarios where loop detection fails or is unavailable. Comparative analysis with methods like UWB-VO confirms the robustness of our approach even with lower measurement frequencies and more realistic operational constraints. Finally, we evaluate and quantify how different geometric arrangements of base stations (Tetrahedral, Z-Shape, Asymmetric, Diamond, and Clustered) affect SLAM performance, demonstrating that optimal configurations can improve localization accuracy.

\subsection{Contributions}
The key contributions of this paper are as follows:
\begin{itemize}
    \item \textbf{Integration of 5G ToA Measurements for Global SLAM}: We introduce a novel approach that integrates 5G \toa measurements into the ORB-SLAM3 framework. This integration anchors the SLAM system's local reference frame to a global coordinate system defined by 5G base stations, enabling accurate and consistent global localization and mapping.
    
    \item \textbf{Resolving Scale Ambiguity in Monocular SLAM}: By incorporating ToA measurements, our method effectively resolves the scale ambiguity inherent in monocular SLAM systems. This allows for accurate metric scale estimation without the need for additional sensors or manual scaling.
    
    \item \textbf{Enhanced Robustness in Challenging Scenarios}: We demonstrate system performance with unknown base station positions and sequential visibility constraints, showing that even partial or intermittent ToA measurements can significantly improve localization accuracy and consistency.
    
    \item \textbf{ToA as an Alternative to Loop Closure}: We evaluate the effectiveness of ToA integration as a substitute for traditional loop closure mechanisms, showing comparable or superior performance in maintaining trajectory consistency, particularly beneficial in environments where loop detection is unreliable or impossible.
    
    \item \textbf{Comprehensive Evaluation Against State-of-the-Art}: We validate our approach using multiple datasets including both custom Aerolab collections and the standard EuRoC MAV benchmark, providing direct comparisons with state-of-the-art methods like UWB-VO under realistic operational constraints.
    
    
\end{itemize}

This paper is organized as follows. Section \ref{sec:related_work} reviews related work in visual SLAM, visual-inertial SLAM, and radio-based localization and SLAM. Section \ref{sec:methodology} details the methodology for integrating ToA measurements into ORB-SLAM3. Section \ref{sec:experiments} describes the experimental setup and datasets used for evaluation. Section \ref{sec:resuls} presents and analyzes the results of the proposed method. Section \ref{sec:limitation} discusses limitations and potential areas for future research, and Section \ref{sec:conclusion} concludes the paper with key findings and implications.

\section{Related Work}
\label{sec:related_work}

We review the literature under three categories: visual SLAM, visual-inertial SLAM, and radio-based SLAM, summarizing key advancements and challenges in each domain.

\subsection{Visual Slam}
Visual SLAM systems, which rely primarily on camera sensors, have undergone significant evolutionary stages in algorithmic development. Feature-based methods pioneered early advancements by extracting keypoints and descriptors from images to estimate camera motion and 3D structure. MonoSLAM ~\cite{davison2007monoslam} introduced real-time monocular SLAM using an Extended Kalman Filter (EKF) to track sparse feature points, laying groundwork for subsequent approaches. Parallel Tracking and Mapping (PTAM) ~\cite{klein2007parallel} revolutionized the field by separating tracking and mapping into parallel threads, enabling more efficient Bundle Adjustment (BA) for improved pose estimation.

The ORB-SLAM series ~\cite{mur2015orb, mur2017orb, campos2021orb} marked a significant milestone by introducing robust feature extraction using Oriented FAST and Rotated BRIEF (ORB) descriptors ~\cite{rublee2011orb}. These implementations progressively expanded SLAM capabilities, with ORB-SLAM2 supporting monocular, stereo, and RGB-D cameras, and ORB-SLAM3 integrating visual-inertial capabilities and multi-map SLAM techniques.

Direct methods represent another critical approach, operating directly on pixel intensities without explicit feature extraction. Dense Tracking and Mapping (DTAM) ~\cite{newcombe2011dtam} was among the first to use an inverse-depth representation for dense map construction. Large-Scale Direct SLAM (LSD-SLAM) ~\cite{engel2014lsd} introduced a semi-dense approach focusing on high-gradient pixels, while Direct Sparse Odometry (DSO) ~\cite{engel2017direct} minimized photometric error over selected pixels.

Hybrid methods emerged to balance the strengths of feature-based and direct approaches. Semi-Direct Visual Odometry (SVO) ~\cite{forster2016svo} utilized direct methods for rotation estimation and feature-based methods for translation, achieving high-speed performance in dynamic environments. Direct Sparse Mapping (DSM) ~\cite{zubizarreta2020direct} further advanced this paradigm by introducing photometric Bundle Adjustment for global optimization.

Despite significant advancements, visual SLAM continues to face critical challenges: scale ambiguity in monocular systems, sensitivity to environmental conditions like lighting and texture variations, and the persistent need to balance computational complexity with real-time performance. 

\subsection{Visual-Inertial SLAM}

Integrating Inertial Measurement Units (IMUs) with cameras addresses some limitations of visual SLAM by providing complementary motion information. IMUs offer high-frequency localization and orientation data, compensating for visual sensor shortcomings in fast motion or low-light conditions. Conversely, visual sensors help mitigate the cumulative drift commonly associated with IMUs, leading to more accurate and robust localization and mapping.

Filter-based approaches, such as Multi-State Constraint Kalman Filter (MSCKF) ~\cite{Mourikis2007}, MSCKF 2.0 ~\cite{Paul2017}, and Robust Visual-Inertial Odometry (ROVIO) ~\cite{Bloesch2015}, utilize recursive estimation techniques to fuse visual and inertial data. These methods primarily use Extended Kalman Filters (EKFs) to track system states and estimate uncertainties. MSCKF introduced a novel feature marginalization technique to decrease computational complexity, showing robustness during aggressive movements and brief feature losses. ROVIO extended this approach by incorporating direct photometric error minimization, particularly improving performance in low-texture environments.
Optimization-based methods, including Open Keyframe-Based Visual-Inertial SLAM (OKVIS) ~\cite{leutenegger2015keyframe}, Visual-Inertial Navigation System (VINS-Mono) ~\cite{qin2018vins}, VINS-Fusion ~\cite{qin2019general}, and ORB-SLAM3 ~\cite{campos2021orb}, solve for system states by minimizing cost functions over data windows. These approaches employ advanced non-linear optimization techniques like bundle adjustment to ensure consistent mapping and localization throughout the trajectory. VINS implementations, for instance, utilized sliding window optimization frameworks that demonstrated remarkable robustness across diverse real-world datasets, while ORB-SLAM3 comprehensively integrated IMU data across multiple camera modalities.

Challenges in visual-inertial SLAM include accurate intrinsic and extrinsic calibration, robust initialization that involves scale and gravity estimation, handling dynamic environments, maintaining computational efficiency, and mitigating the effects of highly noisy IMU data. 

\subsection{Radio-Based Localization and SLAM}

While visual and visual-inertial SLAM methods have shown remarkable performance, they often struggle in feature-sparse environments, under poor lighting conditions. These methods can also suffer from accumulated drift over time and the lack of absolute reference frames, particularly in GPS-denied or indoor environments.

Radio-based SLAM has emerged as a valuable complement to traditional methods. It leverages Radio Frequency (RF) signals—such as Ultra-Wideband (UWB), Wi-Fi, millimeter-wave (mmWave), and 5G technologies—for localization and mapping.

Recent advancements in localization techniques have explored innovative approaches for positioning mobile receivers in challenging environments ~\cite{gentner2016multipath}. Gentner et al. introduced Channel-SLAM, an algorithm leveraging multipath signals for positioning by treating multipath components as virtual transmitters and employing recursive Bayesian filtering with a Rao-Blackwellized particle filter. This approach demonstrates the potential of exploiting reflections and scattering without prior environmental knowledge, achieving accurate positioning in both line-of-sight and non-line-of-sight conditions. However, the method is computationally intensive and limited to multipath signal processing without incorporating visual or inertial sensors.

In a graph-based framework in radio SLAM, researchers have explored belief propagation and factor graphs to jointly localize mobile agents and map environments using multipath components~\cite{leitinger2017factor, leitinger2019belief}. Specular reflections are modeled as virtual anchors (VAs), which are mirror images of physical anchors (PAs), allowing the system to simultaneously estimate the positions of VAs, PAs, and the mobile agent.  Such techniques enhance localization accuracy by incorporating advanced signal parameters like the Angle of Arrival but often face challenges in computational complexity. Similarly, Chu et al. \cite{chu2021vehicle} and Mendrzik et al. \cite{mendrzik2018joint} have investigated joint localization and radio mapping, using multipath information to estimate vehicle positions and environmental features simultaneously via factor graphs. However, their work neither utilized real data nor incorporated realistic network simulations and fusion schemes necessary for practical implementations.

Complementary approaches have utilized Wi-Fi technologies, such as 60 GHz IEEE 802.11ad, to provide indoor localization. Bielsa et al. \cite{bielsa2018indoor} have developed a real-time system that achieves sub-meter accuracy in 70\% of cases. While promising, these Wi-Fi-based methods may be limited by the quality of the wireless signal.

The emergence of 5G technology has further expanded localization possibilities for localization ~\cite{karfakis2023nr5g, ferre2019positioning}. Researchers developed frameworks like NR5G-SAM, presented in \cite{karfakis2023nr5g}, combined ToA and Received Signal Strength Indicator (RSSI) measurements with inertial sensing, employing factor graphs for trajectory estimation and Radio Environmental Map (REM) creation. This approach showed promise, especially in GNSS-denied and rural areas. However, it faced several limitations. The reliance on RSSI-based mapping reduced precision, particularly in vertical positioning. Additionally, the absence of loop closure mechanisms and dependence on multilateration increased computational demands.  Del Peral-Rosado et al.~\cite{PeralRosado2016} and Saleh et al.\cite{Saleh2022} explored 5G-based positioning techniques, achieving accuracies of 20-25 cm and sub-meter level, respectively.  Other 5G-based localization methods have explored various techniques, including fingerprinting, machine learning, and signal processing. Talvitie et al. \cite{talvitie2018positioning} utilized 5G synchronization signals to achieve sub-meter accuracy for high-speed train tracking. Zhang et al. \cite{zhang2021aoa} employed a deep neural network to improve positioning accuracy using 5G AoA and amplitude information, even in non-line-of-sight environments. Shamaei and Kassas \cite{shamaei2021receiver} proposed an opportunistic ToA estimation approach using 5G synchronization signals and PBCH, achieving a ranging error standard deviation of 1.19 m. All of these studies primarily focused on localization and did not address the simultaneous mapping aspect of SLAM.

 Recent research has also explored the integration of 5G Time-of-Arrival (ToA) measurements with inertial data for indoor localization. Kabiri et al. \cite{kabiri2023pose, kabiri2024graph} have investigated graph-based optimization and Error State Kalman Filter (ESKF)-based fusion approaches for Micro Aerial Vehicle (MAV) indoor localization. While these methods demonstrate promising results, they primarily focus on localization and do not explicitly address the simultaneous mapping aspect of SLAM. 

The integration of Ultra-Wideband (UWB) technology with visual SLAM has emerged as another promising approach for improving localization accuracy. Lin and Yeh \cite{lin2022drift} proposed a drift-free visual SLAM technique for mobile robot localization by integrating UWB positioning technology. Their approach utilizes the global constraint of UWB positioning to reduce locally accumulated errors of visual SLAM localization based on the extended Kalman filtering (EKF) framework. Experimental results demonstrated that the integration of UWB positioning reduced the overall drift error of robot navigation by more than 50\%. However, this approach relies on a dense network of UWB anchors and primarily addresses drift reduction rather than providing a comprehensive framework for global localization.

Li et al. \cite{li2023uwb} proposed UWB-VO, a tightly coupled SLAM framework that combines monocular vision with UWB distance measurements. Their method jointly optimizes UWB distance residuals and visual reprojection errors to improve positioning accuracy and address scale ambiguity in monocular SLAM, achieving scale error rates below 1\%. The system features a UWB anchor position optimization technique for rapid initialization and provides comparative benchmarks against several visual-inertial odometry algorithms. However, UWB-VO operates under the assumption of continuous visibility of at least one UWB anchor and requires a relatively high measurement frequency (30 Hz), which may limit its practicality in dynamic or complex environments.

Research on alternatives to traditional loop closure for drift correction represents another important direction in SLAM development. Conventional loop closure mechanisms rely on revisiting previously mapped areas, which may not be feasible in exploration scenarios, time-critical operations, or environments with significant appearance changes. While several approaches have explored external reference sources for drift correction, few have systematically evaluated radio-based measurements as direct substitutes for loop closure. This gap highlights the need for robust SLAM systems that can maintain trajectory consistency even with limited or intermittent external measurements from imprecisely localized reference points.

Despite recent advancements in radio-based localization methods, particularly those leveraging 5G technologies, these approaches typically focus solely on localization without addressing the simultaneous mapping aspect essential for SLAM applications. Furthermore, most radio SLAM methods do not incorporate sensor fusion with visual data, relying instead on complex multipath signal processing alone. This lack of integration with additional sensor modalities limits their robustness and applicability in diverse and dynamic environments.

Our work distinguishes itself by integrating 5G ToA measurements into a state-of-the-art visual-inertial SLAM system, specifically extending ORB-SLAM3. By incorporating ToA data into the SLAM pipeline, we establish global localization within the reference frame defined by 5G base station placements.

To the best of our knowledge, this is the first study to comprehensively evaluate 5G ToA integration with visual-inertial SLAM across both known and unknown base station scenarios. Our approach addresses the limitations of prior methods by:

\begin{itemize}
\item \textbf{Simultaneous Localization and Mapping with Global Reference}: Unlike previous radio-based localization techniques that focus solely on positioning, our method simultaneously performs mapping while providing global localization, enhancing the overall robustness and applicability of SLAM in indoor environments.

\item \textbf{Integration without Compromising Local Accuracy}: We demonstrate that incorporating ToA measurements does not degrade the local estimation accuracy of ORB-SLAM3. This is crucial because it leverages the precision of visual-inertial SLAM while adding the benefits of global positioning.

\item \textbf{Robustness to Unknown Base Station Configurations}: We demonstrate system performance with unknown base station positions and sequential visibility, showing that our approach remains effective even with partial or intermittent ToA measurements—conditions commonly encountered in real-world deployments.

\item \textbf{Alternative to Traditional Loop Closure}: We provide the first systematic evaluation of ToA integration as a substitute for conventional loop closure mechanisms, demonstrating comparable or superior drift correction without requiring revisits to previously mapped areas.

\item \textbf{Practical Implementation with Realistic Constraints}: Unlike approaches such as UWB-VO that require high-frequency measurements (30 Hz) and continuous anchor visibility, our system operates effectively with lower measurement rates (5-10 Hz) and intermittent base station connectivity, making it more practical for real-world deployment.
\end{itemize}

\section{Methodology}
\label{sec:methodology}

This section outlines the technical framework and processes to integrate ToA data into ORB-SLAM3 for globally consistent localization. We describe the distinction between local and global frames, \toa factor formulation, and the system modifications that enable robust pose estimation across diverse datasets.

\subsection{Local Frame vs. Global Frame in SLAM}

In SLAM, the concepts of local and global reference frames are fundamental for accurately representing and interpreting positional data within a mapped environment. A \textbf{local frame}  is usually established at the beginning of the mission (e.g., at the first keyframe or the sensor's starting point). Once set, all subsequent poses, map points, and sensor measurements are calculated and expressed relative to this local coordinate system. While effective for maintaining local consistency, the local frame lacks alignment with a fixed real-world reference, leading to cumulative drift over time.  Additionally, it can limit the system's ability to integrate or compare data from multiple sessions or external sources, which often require a global reference frame for alignment.

In contrast, a \textbf{global frame} is a fixed, absolute reference system tied to known landmarks or infrastructure, such as 5G base stations in our setup. This global frame provides a consistent anchor point independent of the SLAM system’s starting pose. By aligning the local observations to the global frame, the SLAM system can mitigate drift, maintain long-term positional accuracy, and ensure interoperability with other systems or datasets operating in the same global context.

\begin{figure}
    \centering
    \includegraphics[width=\linewidth]{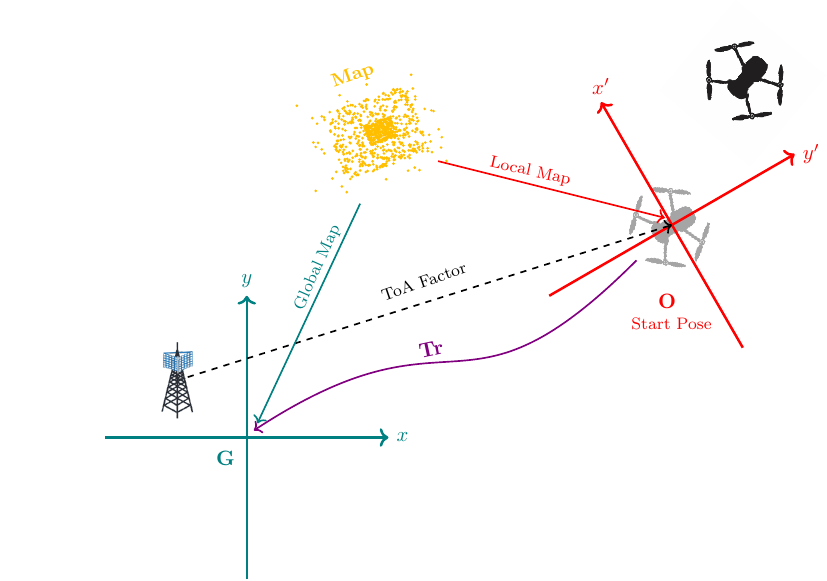}
    \caption[Relationship between local and global frames in SLAM]{Diagram illustrating the relationship between the local and global frames in SLAM. The transformation \( \mathbf{Tr} \) aligns the local frame (\( x', y' \)) with the global coordinate frame (\( x, y \)). The single map, shown in both frames, is represented in red for the local frame and teal for the global frame. The initial pose of the drone is faded to indicate its starting position, while the current pose is shown in the local frame. A base station in the global frame provides \toa measurements, aiding in the alignment and refinement of global positioning. The dashed arrow represents the influence of \toa measurements, linking the local and global frames via \( \mathbf{Tr} \).}
    \label{fig:local_vs_global_map}
\end{figure}

\textbf{Fig.~\ref{fig:local_vs_global_map}} illustrates the distinction between the local and global frames and highlights the transformation process that bridges them. The local map, represented in red, depicts the environment relative to the initial sensor position. However, this map lacks global context, making it unsuitable for applications requiring absolute positioning. Conversely, the global map, depicted in teal, integrates \toa measurements from fixed base stations, which serve as global reference points. The transformation \( Tr \), shown in the figure, aligns the local frame to the global coordinate system,  linking the two maps.

This alignment process is essential for achieving a globally consistent SLAM system. The transformation \( Tr \) is estimated by leveraging ToA measurements, which provide spatial constraints based on the distances between the robot and the base stations. These constraints refine the local-to-global alignment and ensure that the trajectory and map points are accurately represented within the global frame. The dashed arrow in Fig.~\ref{fig:local_vs_global_map} illustrates the influence of ToA measurements in bridging the local and global scales.

The integration of a global reference frame offers several advantages:
\begin{itemize}
    \item \textbf{Drift Mitigation}: Periodic alignment with the global frame reduces cumulative drift inherent in local-only SLAM systems.
    \item \textbf{Multi-Agent Collaboration}: By operating within the same global frame, multiple robots or drones can share maps and coordinate actions effectively.
    \item \textbf{Robustness in Challenging Environments}: In scenarios with limited visual or inertial cues (e.g., featureless or dynamic environments), the global reference frame, established through methods such as 5G base stations or GPS, provides consistent positional information, ensuring robustness and accuracy.
\end{itemize}

This distinction between local and global frames underpins the methodology used in this work, particularly in the integration of ToA data within ORB-SLAM3. 

\subsection{ToA Factor Formulation}

\begin{figure}[h]
    \centering
    \includegraphics[width=0.5\linewidth]{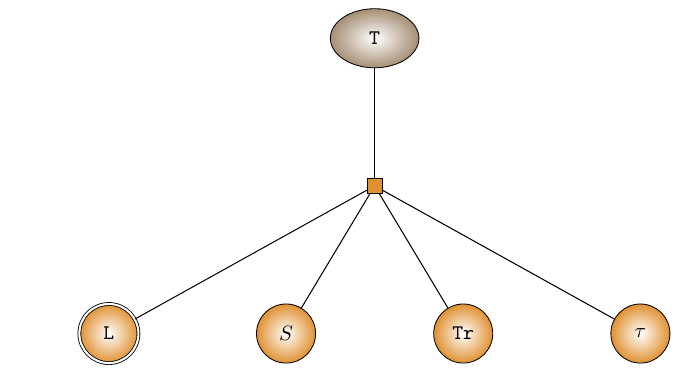} 
    \caption[Structure of the ToA factor]{Structure of the ToA factor, illustrating the key components: camera pose node (\( \mathrm{T} \)), scale factor (\( s \)), local-to-global transformation node (\( \mathrm{Tr} \)), base station position node ($\mathrm{L}$), and bias node (\( \tau \)). Double-bordered nodes indicate fixed parameters during optimization.}
    \label{fig:ToA_factor_structure}
\end{figure}

\toa factor integrates distance measurements from base stations into the SLAM framework through a formulated error model. The factor connects multiple optimization vertices: the camera pose, local-to-global transformation, ToA measurement bias, and, in monocular cases, a scale factor. Fig.~\ref{fig:ToA_factor_structure} illustrates the structure of the ToA factor, highlighting how each component interacts within the optimization graph. The double-bordered base station nodes indicate that these parameters are fixed and their positions are known.

\subsubsection{Mathematical Formulation}

Let $d_\text{meas}$ represent the measured distance affected by noise $\noise^\text{Dist}$ and bias $\tau$:


\begin{equation}
    d_\text{meas} = d_\text{true} + \noise^\text{Dist} + \tau
\end{equation}

The factor involves the following key components:
\begin{itemize}
    \item Camera pose in the SLAM frame (local frame): $\mathbf{T}_{oc}$
    \item Local-to-global transformation: $\mathbf{T}_{go}$
    \item Base station position in global frame: $L_\text{G}$
    \item Scale factor (for monocular systems): $s$
    \item Bias of ToA distance measurement: $\tau$  
\end{itemize}

The transformation chain to calculate the keyframe pose in the global frame is given by:
\begin{equation}
    \mathbf{T}_{gc} = \mathbf{T}_{go} \cdot \mathbf{T}_{oc}
\end{equation}

The calculated distance between the camera and the base station is:
\begin{equation}
    d_\text{calculated} = ||s \cdot t_\text{gc} - L_\text{G}||_2
\end{equation}

where $t_\text{gc}$ is the translation component of $\mathbf{T}_{gc}$, representing the camera's position in the global frame and $s$  is the scaling parameter applied to account for scale differences, crucial in monocular SLAM systems where scale ambiguity exists. By introducing \( s \) into the error computation, the optimization can estimate the true metric scale of the environment:

\begin{itemize}
    \item If \( s \) is known (e.g., in stereo or RGB-D systems), it can be fixed at \( s = 1 \).
    \item In monocular systems, \( s \) becomes an additional variable to estimate during optimization.
\end{itemize}

The ToA factor error is then computed as:
\begin{equation}
    e = d_\text{calculated} - (d_\text{meas} - \tau ) 
\end{equation}

The optimization framework minimizes a combined cost function for \toa measurements. For a single base station, the objective is to minimize the squared error between the measured and estimated TOAs:
\begin{equation}
    \min_{\mathbf{T}_{oc}, \mathbf{T}_{go}, \tau, s} \sum_{i} (e_i)^2
\end{equation}

where $i$ indexes the individual \toa measurement, and $e_i$ is the error between the measured and estimated TOA for the i-th measurement. For multiple base stations, the objective function is extended to minimize the sum of squared errors across all base stations.

\begin{equation}
\min_{ \mathbf{T}_{oc}, \mathbf{T}_{go}, \tau_j, s} \sum_{i} \sum_{j} (e_{i,j})^2
\end{equation}
where the summation spans all ToA measurements and their corresponding optimization variables.

\begin{itemize}
\item $j$ indexes the base stations
\item $\tau_j$ is the bias for the $j$-th base station
\item $e_{i,j}$ is the error between the measured and estimated \toa for the $i$-th measurement at the $j$-th base station
\end{itemize}

\subsubsection{Uncertainty Propagation}
The information matrices (inverse covariance matrices) of both the scale parameter and the local-to-global transformation are suitably updated through the optimization process. These matrices are computed using the Hessian approximation derived from the Jacobians of the error terms with respect to the corresponding vertices. The updated uncertainties serve as prior information in subsequent optimization iterations, enabling a more robust convergence by:
\begin{itemize}
    \item Providing appropriate weighting for new measurements based on accumulated certainty
    \item Preventing aggressive updates based on noisy or conflicting measurements to well-established estimates
    \item Allowing faster adaptation when uncertainty is high
\end{itemize}




\subsection{System Components and Integration}

\subsubsection{System Overview}

The proposed system integrates \toa measurements into the ORB-SLAM3 framework to achieve globally consistent localization and mapping. The main components of the system include tracking, local mapping, loop closing \& ToA-based global map refinement. These components interact in a multi-threaded architecture to ensure real-time performance. 

\begin{figure}
    \centering
    \includegraphics[width=1\linewidth]{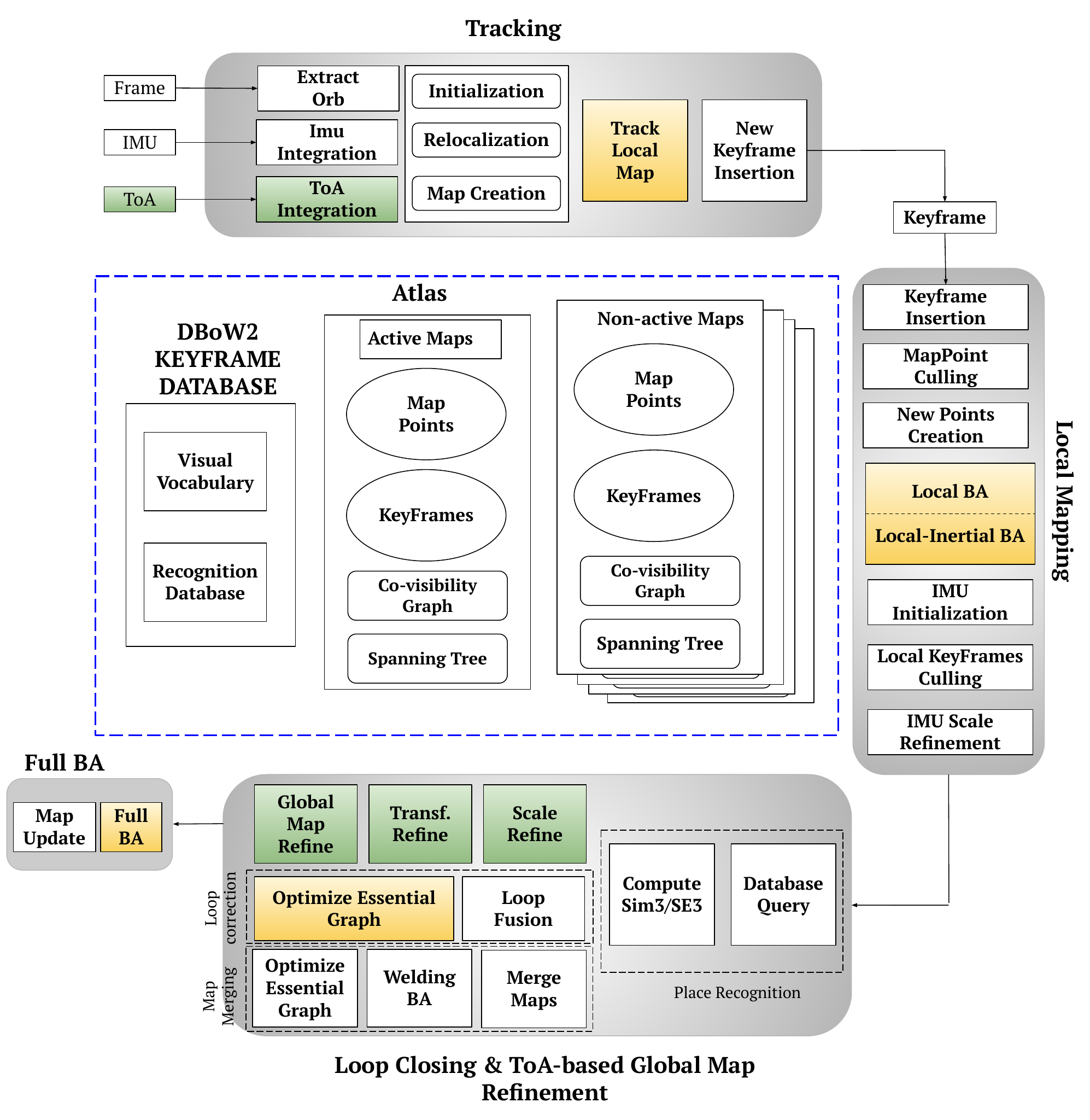}
    \caption[ORB-SLAM3 pipeline with ToA integration]{Diagram of ORB-SLAM3 with ToA integration, illustrating the pipeline across Tracking, Local Mapping, and Loop Closing threads. New components are shown in green, including ToA-based global map refinement, while modified components are in yellow, such as Local Bundle Adjustment.}
    \label{fig:orb_diag}
\end{figure}

Fig.~\ref{fig:orb_diag} illustrates the system architecture, highlighting how ToA measurements are incorporated into the SLAM pipeline. Several existing components have been modified, including the Local Bundle Adjustment module, which now incorporates ToA measurements for improved optimization accuracy. The system diagram uses color coding to distinguish between modifications to the original ORB-SLAM3 framework: new components are highlighted in green, while modified components are shown in yellow.

\subsubsection{System Integration}

The ToA measurements are integrated at multiple levels within the ORB-SLAM3 framework to maximize their utility while maintaining real-time performance, i.e., Tracking, Local Mapping, Loop Closing-ToA-based global map refinement. In what follows, we highlight some of the key changes and additions. 

\begin{figure*}[htbp]
    \centering
    \resizebox{0.7\textwidth}{!}{%
    \begin{minipage}{\textwidth}
        \begin{subfigure}{\linewidth}
            \centering
            \includegraphics[width=1\linewidth]{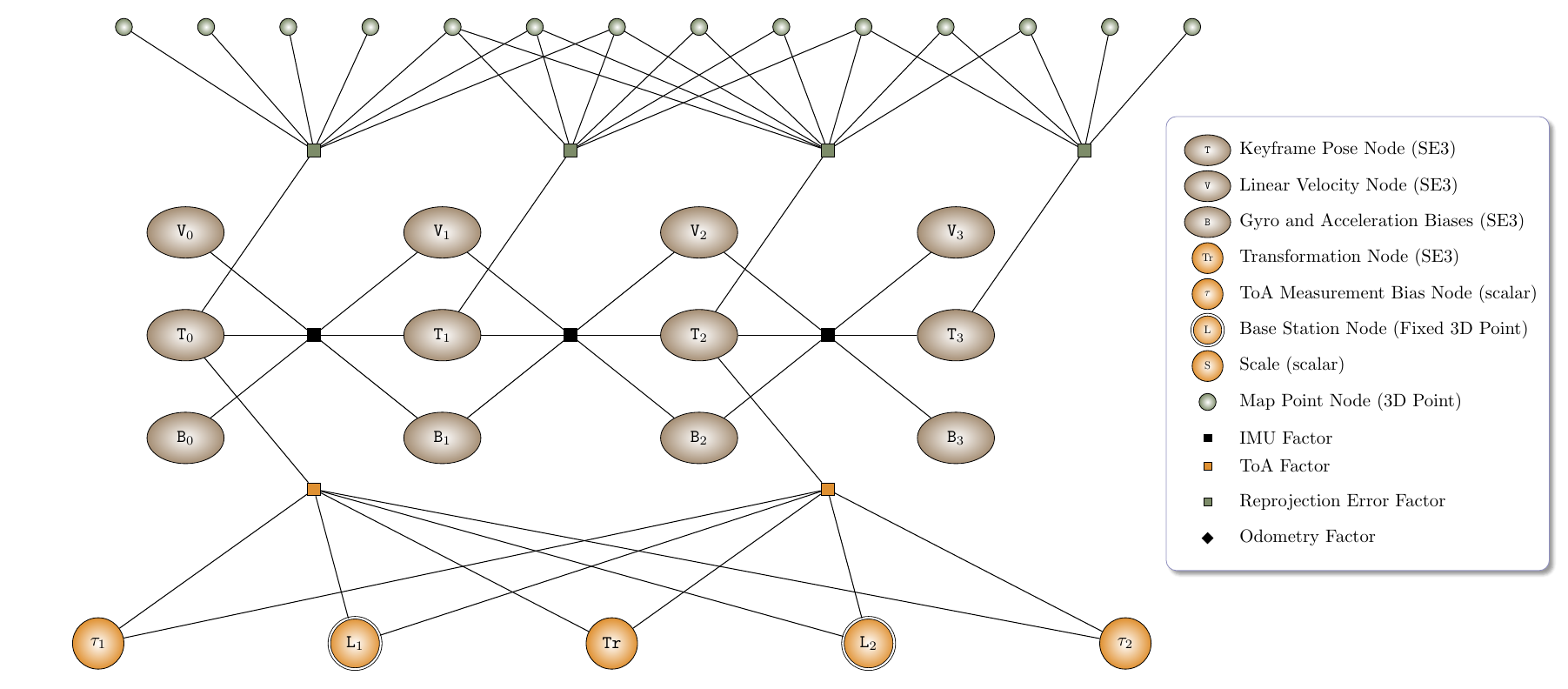}
            \caption{Local Inertial BA}
            \label{fig:graph_localinertialBA}
        \end{subfigure}
        
        \vspace{0.5cm} 
        
        \begin{subfigure}{0.48\linewidth}
            \centering
            \includegraphics[width=\linewidth]{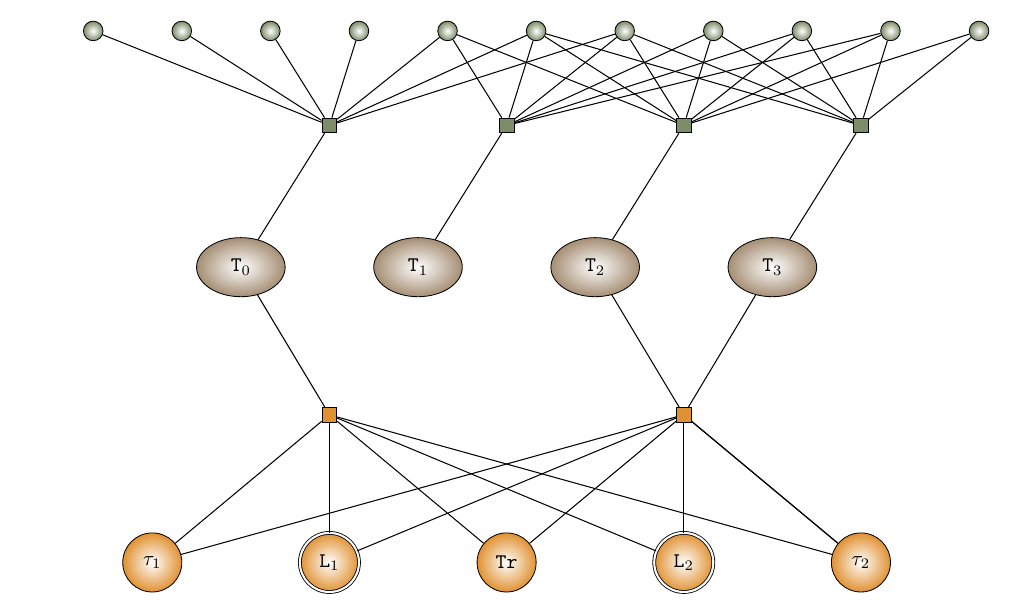}
            \caption{Local Bundle Adjustment}
            \label{fig:graph_BA}
        \end{subfigure}
        \hfill
        \begin{subfigure}{0.48\linewidth}
            \centering
            \includegraphics[width=\linewidth]{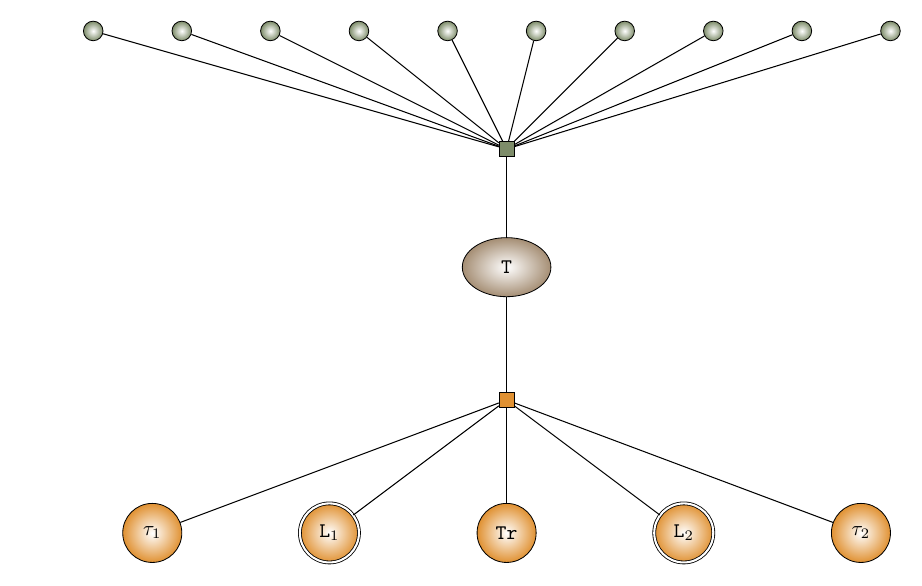}
            \caption{Tracking Pose Optimization}
            \label{fig:graph_po}
        \end{subfigure}
        
        \vspace{0.5cm} 
        
        \begin{subfigure}{0.48\linewidth}
            \centering
            \includegraphics[width=\linewidth]{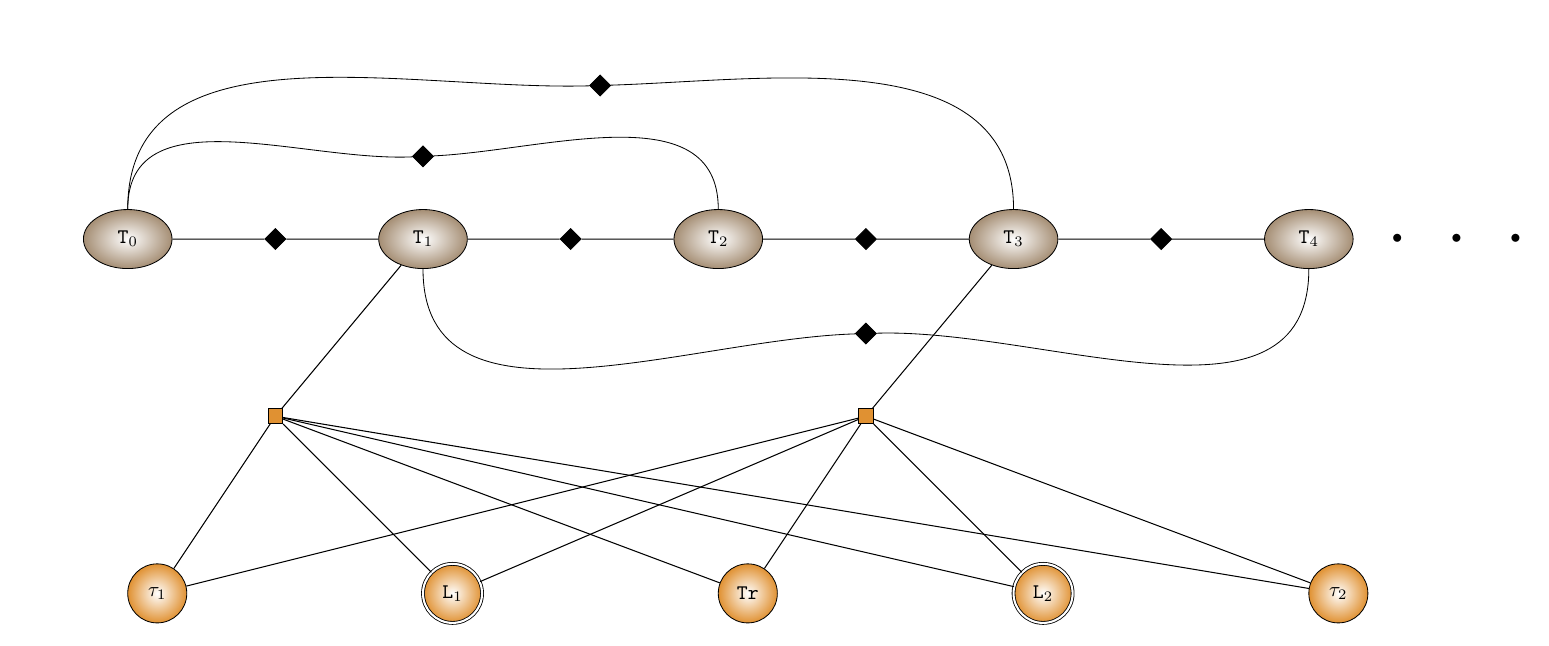}
            \caption{Global Map Refinement}
            \label{fig:graph_globmapref}
        \end{subfigure}
        \hfill
        \begin{subfigure}{0.48\linewidth}
            \centering
            \includegraphics[width=\linewidth]{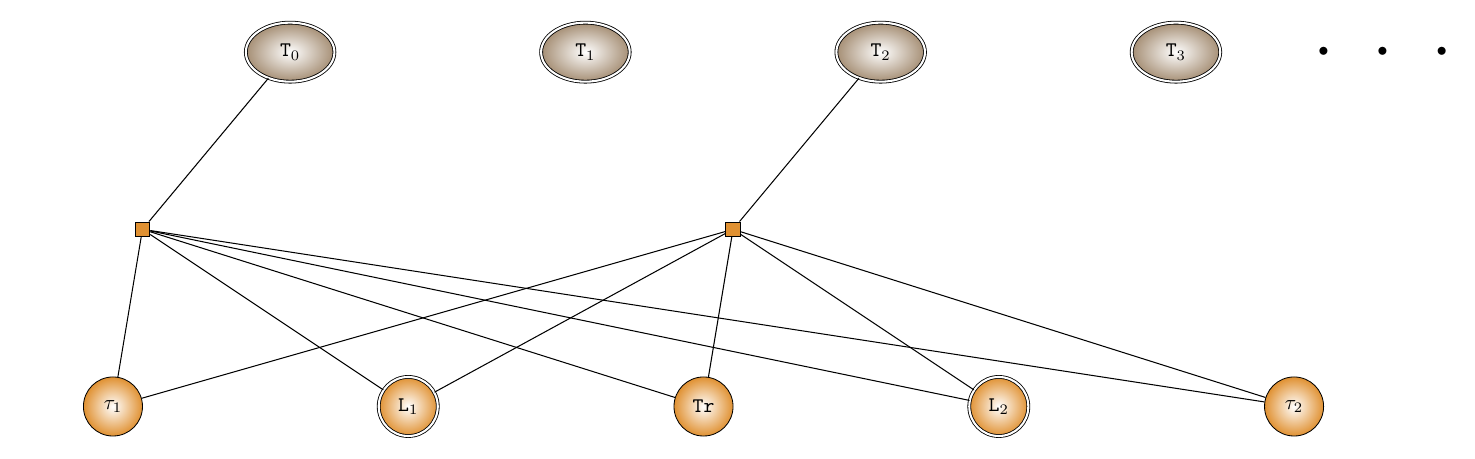}
            \caption{Transformation Refinement}
            \label{fig:transformation_refinement}
        \end{subfigure}
        
        \vspace{0.5cm} 
        
        \begin{subfigure}{\linewidth}
            \centering
            \includegraphics[width=0.4\linewidth]{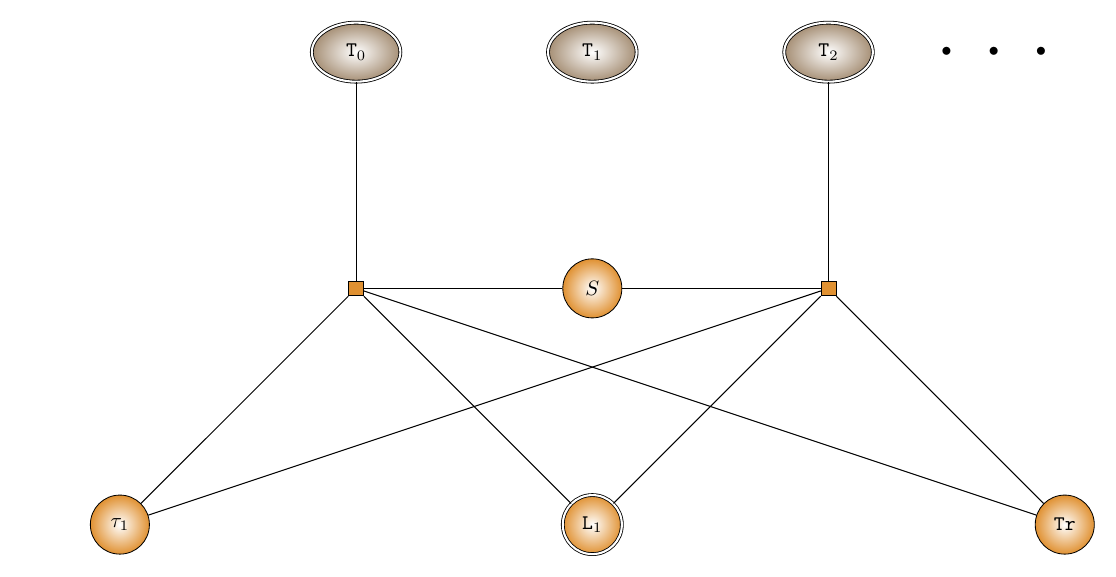} 
            \caption{Scale Refinement}
            \label{fig:scale_graph}
        \end{subfigure}
    \end{minipage}
    }

    \caption [SLAM optimization graph structures for modified and new components]{
    Structures of optimization graphs for various components in the SLAM process:
    \textbf{(a) Local Inertial BA}: Optimization that integrates visual, inertial, and ToA measurements to refine keyframe poses and map points while addressing significant IMU noise. 
    \textbf{(b) Local Bundle Adjustment}: Local refinement of keyframe poses, map points, \toa biases, and local-to-global transformations using visual and ToA constraints within a local optimization window.
    \textbf{(c) Tracking Pose Optimization}: Real-time optimization of the camera pose for the incoming frame by minimizing reprojection errors and incorporating ToA distance constraints.
    \textbf{(d) Global Map Refinement}: Global map refinement by leveraging odometry, co-visibility, loop closure, and ToA edges to improve global map consistency and keyframe accuracy.
    \textbf{(f) Transformation Refinement}: Periodic optimization focusing solely on refining the local-to-global transformation with fixed keyframe poses, enhancing robustness against IMU noise and drift.
    \textbf{(e) Scale Refinement}: A dedicated process for monocular SLAM systems to resolve scale ambiguity by optimizing the global scale factor, ensuring consistent keyframe poses and map points.

    \textbf{Note}: Double-bordered nodes in the graphs indicate fixed nodes during optimization.
    }

    \label{fig:composite_figure}
\end{figure*}

\paragraph{Tracking Thread}
\toa factors are incorporated into frame-to-frame pose optimization, providing additional constraints during tracking. While this integration offers modest improvements, it can enhance robustness during rapid motion or feature-poor sequences. Fig. \ref{fig:graph_po} illustrates the optimization structure in the Tracking thread, showing the interconnected nodes for the coming frame pose, map points, the local-to-global transformation, ToA biases, and base station positions (two are shown in the example). Two main factor types are integrated into the optimization process: ToA distance factors, as discussed earlier, and reprojection error factors. Reprojection error factors assess the difference between the observed location of a map point in an image and its projected position based on the estimated camera pose. By minimizing this error, we optimize the camera pose's alignment with the map.

\paragraph{Local Mapping Thread}
Beyond the Tracking thread, ToA data is integrated into the Local Mapping thread to refine keyframe poses, map points, and the local-to-global transformation. This is achieved through two optimization components:

\begin{itemize}
    \item For systems without IMU data, the Local Bundle Adjustment incorporates visual features alongside the ToA measurements to optimize the poses of keyframes within the local optimization window, as well as the associated map points and the local-to-global transformation. This is illustrated in Fig. \ref{fig:graph_BA}, where the optimization graph includes nodes for keyframe poses, map points, and the transformation between the local and global coordinate frames.
    
   \item Local-Inertial Bundle Adjustment: For systems equipped with an IMU, the Local-Inertial Bundle Adjustment jointly considers visual, inertial, and ToA measurements in the optimization process. This combined optimization leads to more accurate pose and map estimation, especially in the presence of significant IMU noise. The structure of this optimization graph is shown in Fig. \ref{fig:graph_localinertialBA}, where the additional IMU-related factors are incorporated alongside the visual and ToA-based factors.
\end{itemize}

By leveraging ToA measurements in these Local Mapping optimizations, the ORB-SLAM3 system can further refine the estimates of keyframe poses, map points, and the local-to-global transformation, leading to improved overall accuracy and robustness. The inclusion of ToA data is particularly beneficial in scenarios where the visual information alone may be insufficient, such as in the presence of significant IMU noise or in feature-poor environments.

\paragraph{Loop Closing \& ToA-based Global Map Refinement}

One key component in Loop closing is the optimization of the Essential Graph which is responsible for Loop correction in which ToA edges are also added. The Essential Graph is a subset of the full map, containing all the keyframes and four types of edges:

\begin{itemize}
\item Odometry edges: These edges connect consecutive keyframes, representing the relative pose change between them.
\item Covisibility edges: These edges connect keyframes that share a significant number of map points, representing the visual constraints between them.
\item Loop closure edges: These edges are added when a loop closure is detected, providing additional constraints to correct drift and maintain global consistency.
\item ToA-related edges
\end{itemize}
By incorporating ToA measurements into the Essential Graph optimization, we aim to improve the accuracy of keyframe poses by leveraging the additional spatial information from the ToA data and  refine the local-to-global transformation.

\paragraph{New ToA-based Components}

In addition to these modifications, there are also new components that are added, which we explain in the following section.

\textbf{Global Map Refinement:} A periodic global optimization process is performed to maintain the global consistency of the map using \toa. The optimization process is triggered by multiple conditions:
\begin{itemize}
    \item Excessive ToA distance errors beyond measurement covariance thresholds
    \item Significant accumulated motion since last optimization
    \item Time-based triggers ensuring regular refinement
    \item Keyframe count thresholds
\end{itemize}
 Fig.~\ref{fig:graph_globmapref} illustrates the optimization graph structure. It comprises keyframe pose nodes, transformation nodes, and bias nodes, interconnected by three types of edge types, Odometry edges (straight lines), Covisibility edges (angled lines connecting from above or below), and ToA-related edges. 

This integrated approach combines odometry, visual, and ToA constraints to improve the accuracy of the estimated keyframe poses. After the keyframe poses are updated, the map points are also updated accordingly to maintain the consistency of the global map.

\textbf{Transformation Refinement for Inertial Systems:}
In systems equipped with IMU, the initial estimation of the local-to-global transformation can be particularly challenging. When the initial guess for this transformation is far from the true value, the local inertial optimization in the Local Mapping thread can struggle to effectively leverage the ToA measurements to converge to the correct transformation.
To mitigate this issue, we include a dedicated optimization process that focuses solely on refining the local-to-global transformation while keeping the keyframe poses fixed. This approach offers several advantages:
\begin{itemize}

\item Periodic Optimization of the local-to-global Transformation: The transformation refinement optimization is performed periodically within the Loop Closing  \& ToA-based Global Map Refinement thread, ensuring that the local-to-global transformation is regularly updated and maintained.

\item Fixed Keyframe Poses: By keeping the keyframe poses fixed during this optimization, the system can ensure a stable convergence of the transformation estimates, without the additional complexity of jointly optimizing the poses and transformation.

\item Enhanced Robustness against IMU Noise and Drift: The dedicated transformation refinement optimization, with fixed keyframe poses, is more robust against the effects of IMU noise and drift. This is particularly important in scenarios where the IMU measurements contain significant errors, as the optimization can focus solely on refining the transformation without being overly influenced by noisy inertial data.
\end{itemize}
The optimization graph for this transformation refinement process maintains a straightforward structure, incorporating only ToA and odometer edges. Fig.~\ref{fig:transformation_refinement} illustrates this graph structure.

\textbf{Scale Refinement for Monocular Systems:} Monocular visual SLAM systems inherently suffer from scale ambiguity. To address this, we incorporated a dedicated optimization component for global scale factor estimation and update (Fig.~\ref{fig:scale_graph}).

This scale refinement is performed in addition to the local-to-global transformation while keeping keyframe poses fixed. By simplifying the graph to include only ToA edges, we enable global optimization, leveraging information from the entire keyframe map for a more accurate scale estimate. Fig.~\ref{fig:scale_graph} illustrates the structure of the scale refinement optimization graph. Key elements include: Keyframe pose nodes, scale factor node (s), ToA biases, and the Transformation node.

Optimizing the scale factor while keeping keyframe poses fixed efficiently estimates the correct global scale without introducing additional uncertainties or instabilities. The estimated scale factor is then propagated to update all keyframe poses and map point positions, ensuring a consistent and accurate global map representation.

The dedicated scale refinement optimization plays a crucial role in addressing the scale ambiguity inherent to monocular SLAM. While one might consider using the local mapping thread for this purpose, such an approach would be suboptimal for two key reasons. First, the optimization needs to handle significantly higher uncertainty levels, which can lead to instability when processed in local mapping. Second, updating both transformation and scale parameters through local mapping is inefficient, as it operates on a limited map section rather than leveraging the global map information available.

From an implementation standpoint, we utilized the existing three threads within ORB-SLAM3 without creating a new one. Specifically, we integrated the ToA-related global optimizations into the Loop Closing thread, which proved to be a natural and effective choice.

The integration of ToA measurements into the Loop Closing thread was selected for several reasons:

\begin{enumerate}

\item Efficient Asynchronous Execution without Impacting Real-Time Tracking: By incorporating the ToA-based optimizations into the asynchronous Loop Closing thread, we ensure that computationally intensive global optimizations are performed without affecting the real-time tracking performance in the Tracking thread. This approach efficiently utilizes system resources, preventing other threads from being burdened and maintaining overall system performance.

\item Natural Integration with Global Map Maintenance and Separation of Concerns: The Loop Closing thread is responsible for global map consistency tasks like loop closure detection and correction. Integrating the ToA-based optimizations into this thread allows us to seamlessly combine global map refinement processes, leveraging existing infrastructure and workflows. This also maintains a clear separation between local and global optimization tasks—the Tracking thread focuses on local pose optimization, and the Local Mapping thread handles local keyframe and map point refinement—ensuring that each thread operates effectively without interference.

\end{enumerate}

\subsection{System Operation with Unknown Base Station Positions}
While our primary implementation assumes known base station locations to enable global positioning, we also investigate scenarios where these positions are unknown. In this configuration, we modify our optimization framework by changing the base station nodes from fixed to non-fixed status while maintaining a fixed transformation node (TR). This adjustment effectively eliminates the requirement for prior knowledge of base station locations.

The modified optimization graph structure still incorporates ToA measurements, but instead of using them to align the local frame with a globally defined coordinate system, they now serve primarily to constrain the relative geometry of the trajectory and enhance local consistency. While this configuration precludes absolute global positioning capabilities, it remains valuable for improving SLAM performance, particularly in minimal sensor setups where scale ambiguity and drift represent significant challenges.

\section{Experiment}
\label{sec:experiments}
This section presents a detailed evaluation of the proposed methodology, assessing SLAM performance under diverse configurations with simulated 5G ToA measurements.

\subsection{Experimental Setup and Data Collection}
\begin{figure*}[!h]
    \centering
    \begin{subfigure}[t]{0.19\linewidth}
        \centering
        \includegraphics[width=\linewidth]{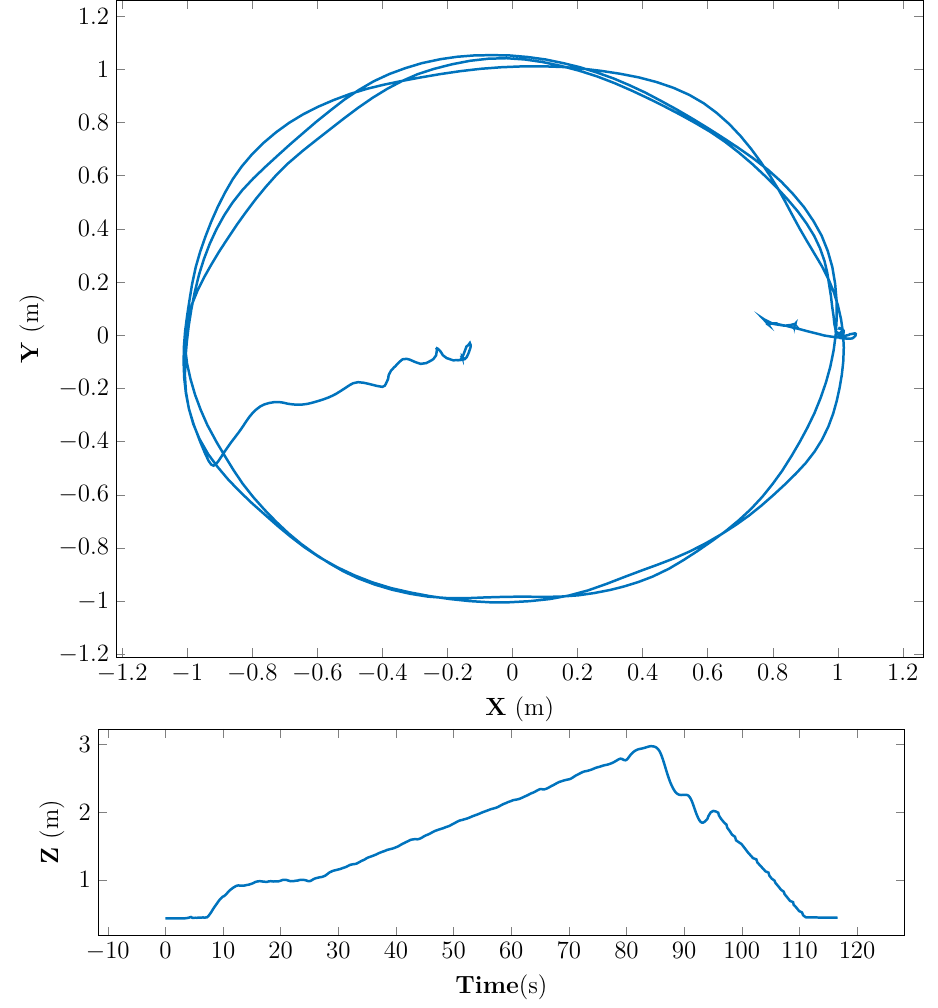}
        \caption{AL01}
        \label{fig:gt0}
    \end{subfigure}
    \hfill
    \begin{subfigure}[t]{0.19\linewidth}
        \centering
        \includegraphics[width=\linewidth]{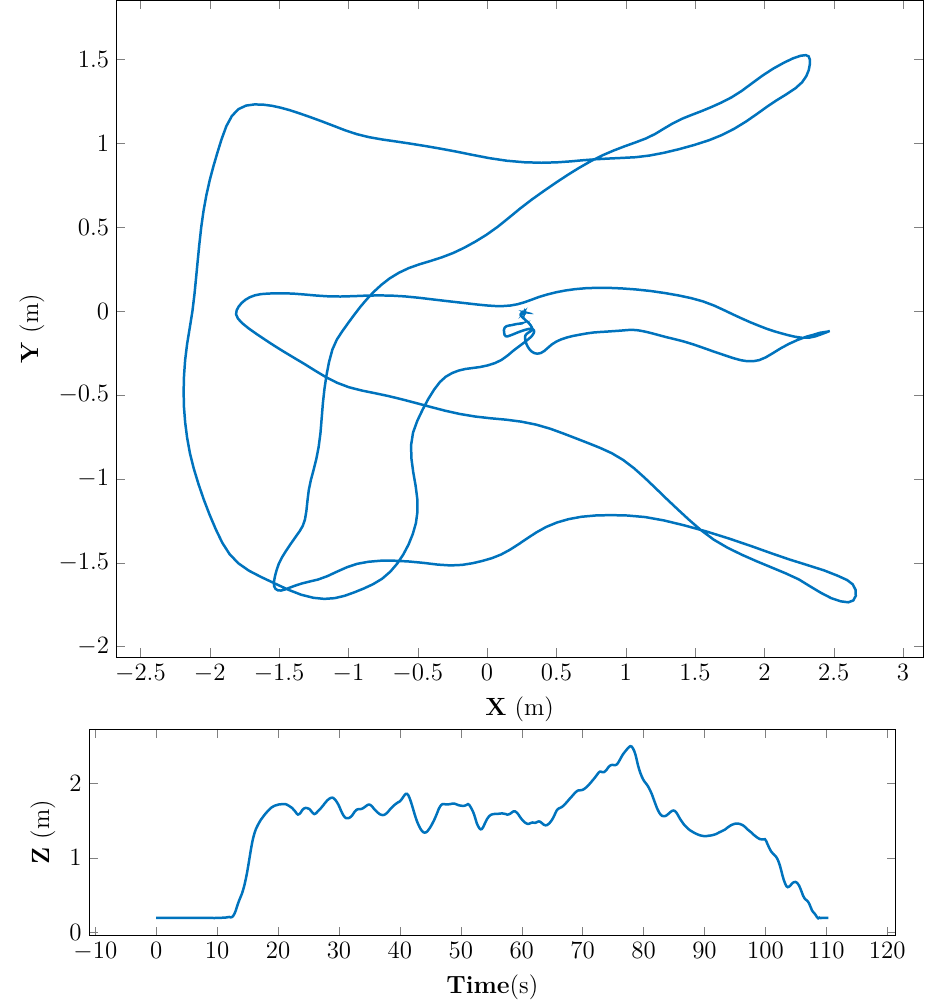}
        \caption{AL02}
        \label{fig:gt1}
    \end{subfigure}
    \hfill
    \begin{subfigure}[t]{0.19\linewidth}
        \centering
        \includegraphics[width=\linewidth]{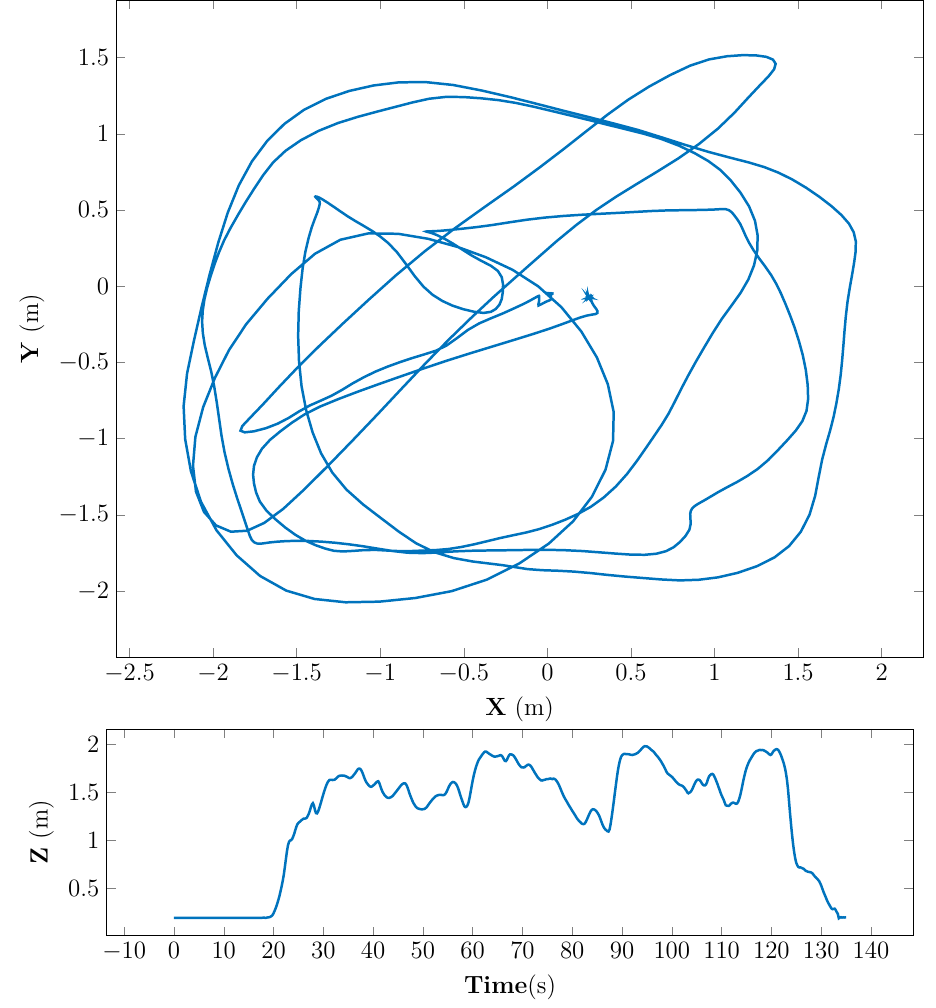}
        \caption{AL03}
        \label{fig:gt2}
    \end{subfigure}
    \hfill
    \begin{subfigure}[t]{0.19\linewidth}
        \centering
        \includegraphics[width=\linewidth]{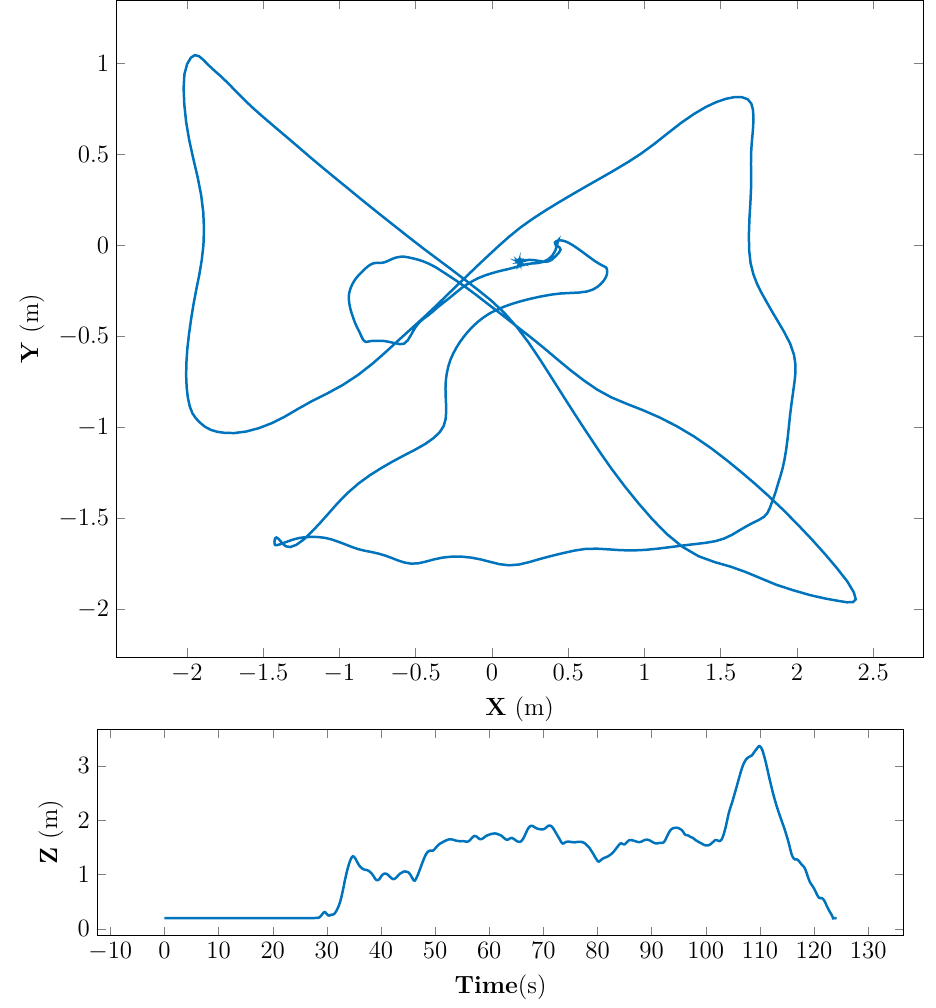}
        \caption{AL04}
        \label{fig:gt3}
    \end{subfigure}
    \hfill
    \begin{subfigure}[t]{0.19\linewidth}
        \centering
        \includegraphics[width=\linewidth]{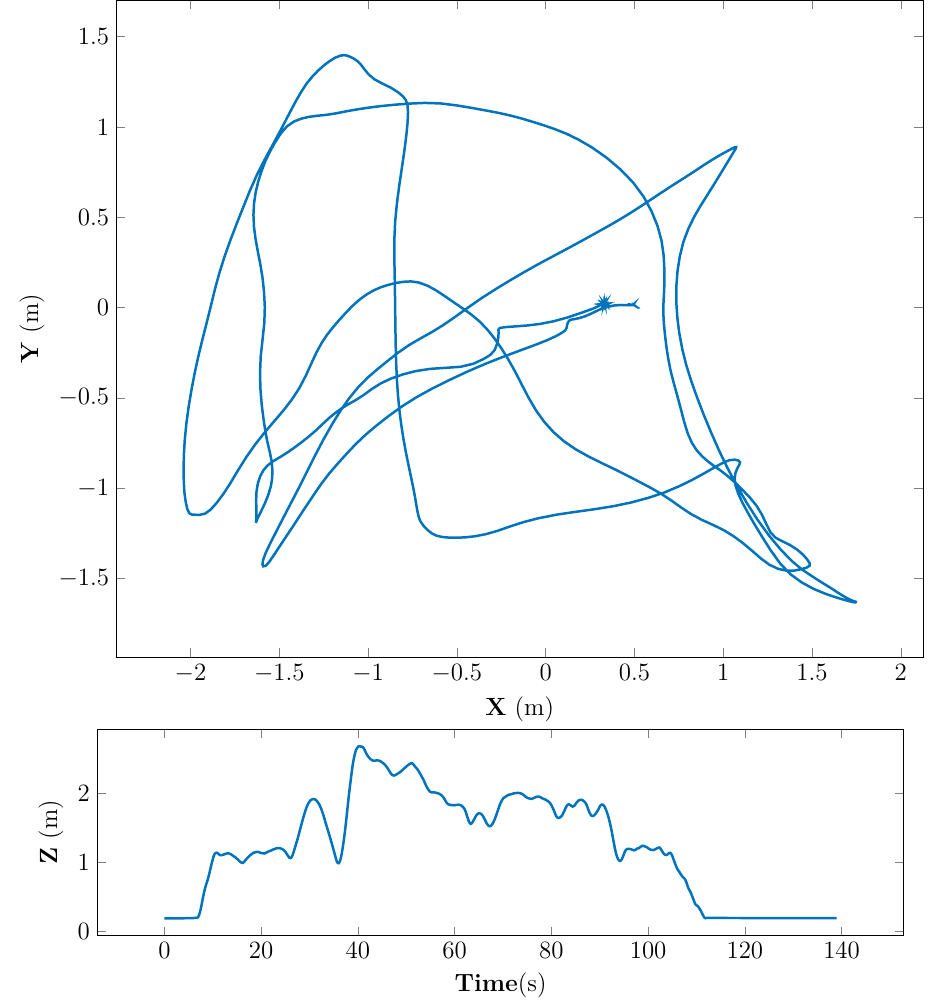}
        \caption{AL05}
        \label{fig:gt4}
    \end{subfigure}
    \caption[Trajectories for the five AeroLab datasets]{Trajectories for the five Aerolab datasets. Each subfigure contains the x-y trajectory (top) and the time vs. z trajectory (bottom) for a specific dataset.}
    \label{fig:gt_traj}
\end{figure*}

\begin{figure*}[h]
    \centering
    \begin{subfigure}[t]{0.32\linewidth}
        \centering
        \includegraphics[width=\linewidth]{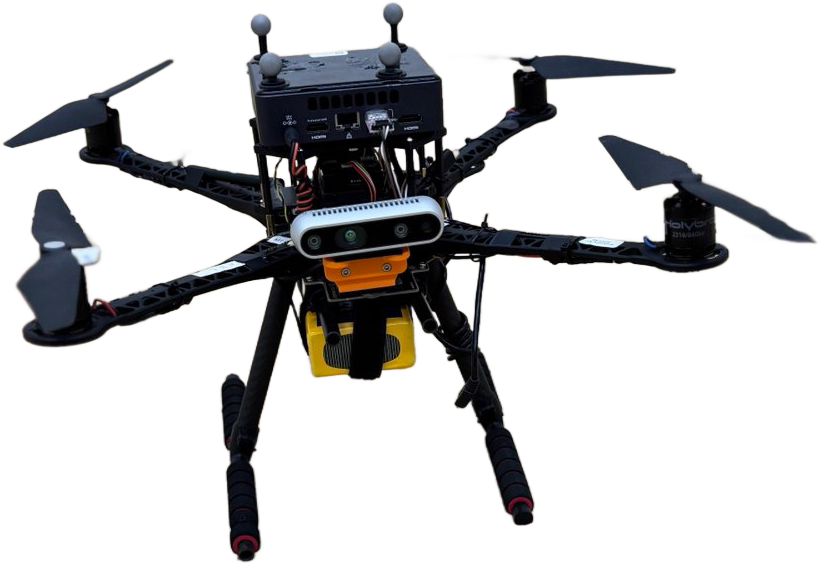}
        \caption{Drone Setup}
        \label{fig:drone2}
    \end{subfigure}
    \hfill
    \begin{subfigure}[t]{0.32\linewidth}
        \centering
        \includegraphics[width=\linewidth]{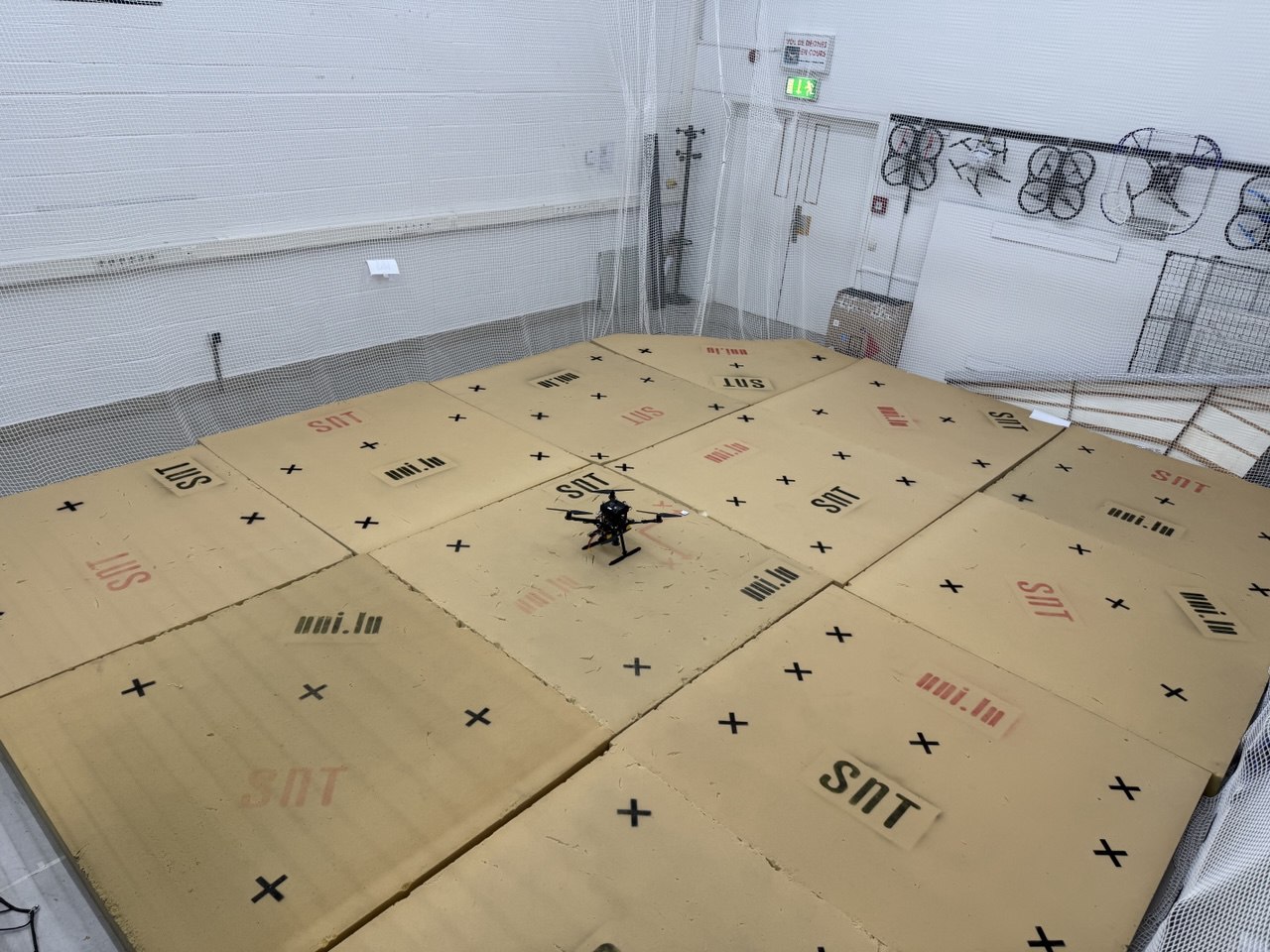}
        \caption{Flight area}
        \label{fig:aero_field}
    \end{subfigure}
    \hfill
    \begin{subfigure}[t]{0.32\linewidth}
        \centering
        \includegraphics[width=\linewidth]{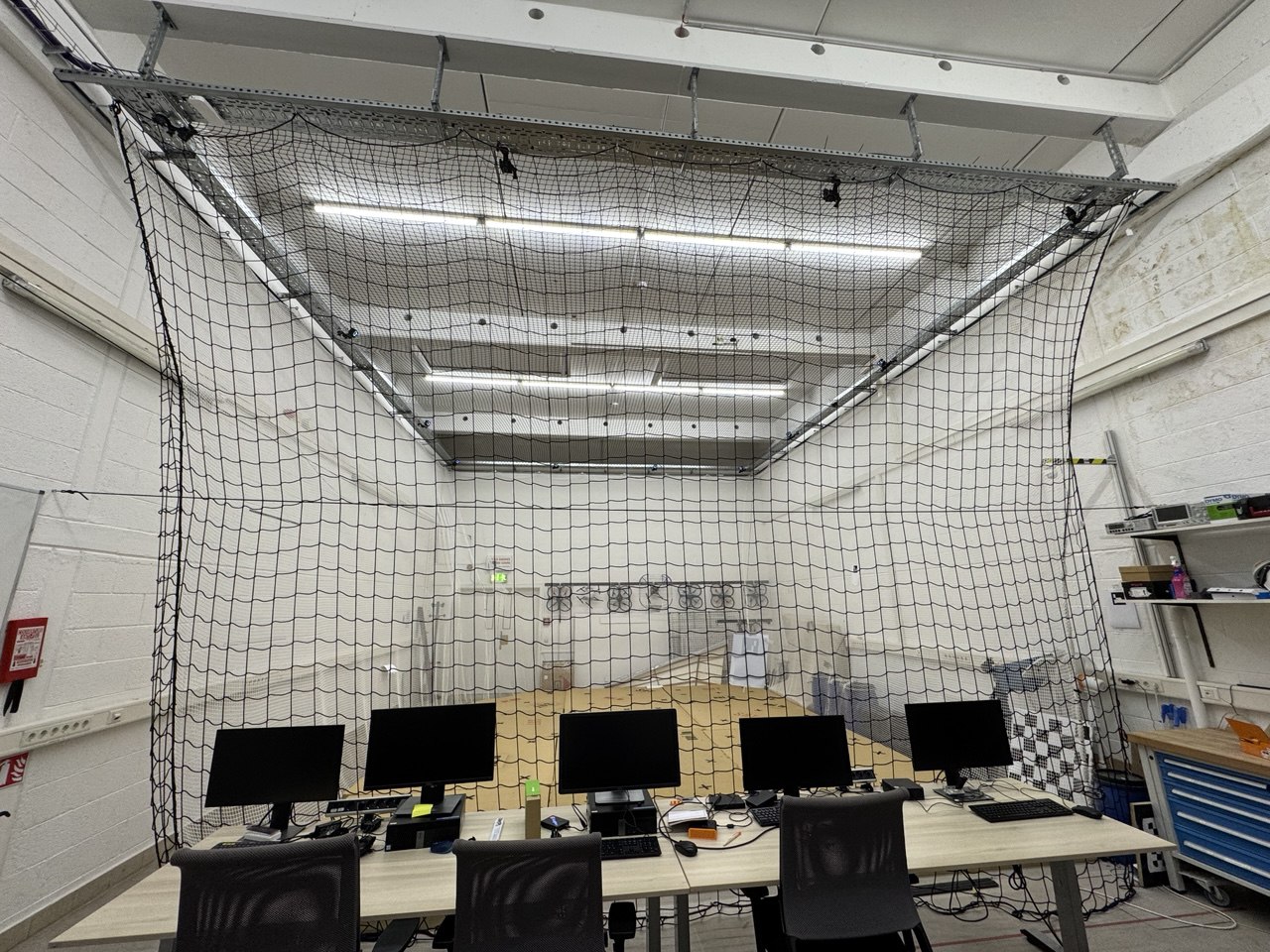}
        \caption{Aerolab overview}
        \label{fig:aero_full}
    \end{subfigure}
    \caption[Drone setup, flight area, and Aerolab overview]{(a) Drone equipped with Intel RealSense D435i camera. (b) Experimental flight area within netted environment. (c) Overall Aerolab view.}
    \label{fig:three_images}
\end{figure*}

The experiments were conducted using two complementary datasets:

\textbf{Aerolab Datasets:} Our custom datasets were collected in the Aerolab indoor environment, designed for drone flights under realistic conditions. We used an Intel RealSense D435i camera for RGB-D (30 Hz) and IMU (200 Hz) data acquisition, with an OptiTrack motion capture system (12 cameras at 120 Hz) providing high-precision ground truth. Five unique datasets were collected through varied flight trajectories ranging from 110-139 seconds Fig.~\ref{fig:gt_traj}. The drone operated within a netted 5x5x5 meter flight area, as shown in Fig.~\ref{fig:three_images}. Calibration of the RealSense camera and IMU was performed using the \texttt{Kalibr} toolbox and \texttt{allan\_variance\_ros}, with data streams automatically synchronized. The simulated ToA measurements were time-aligned with the collected data based on the OptiTrack trajectories.

\textbf{EuRoC MAV Dataset:} To enable comparison with state-of-the-art methods, we also evaluated our approach on the public EuRoC MAV benchmark \cite{burri2016euroc}. This dataset provides stereo images, IMU measurements, and Vicon motion capture ground truth across various indoor environments with different complexity levels. For the Machine Hall sequences (MH01-MH05), we used simple Gaussian ToA simulation to match the conditions in UWB-VO experiments. For the Vicon Room sequences (V101-V203), we employed our comprehensive 5G ToA simulation methodology as detailed in our previous work \cite{kabiri2023pose, kabiri2024graph}.

The experiments were performed on a Ubuntu 20.04 laptop with an Intel i9-10885H CPU and 32 GB of RAM. 

\subsection{5G ToA Simulation and Data Integration}
\begin{figure*}[!h]
    \centering
    \includegraphics[width=0.7\linewidth]{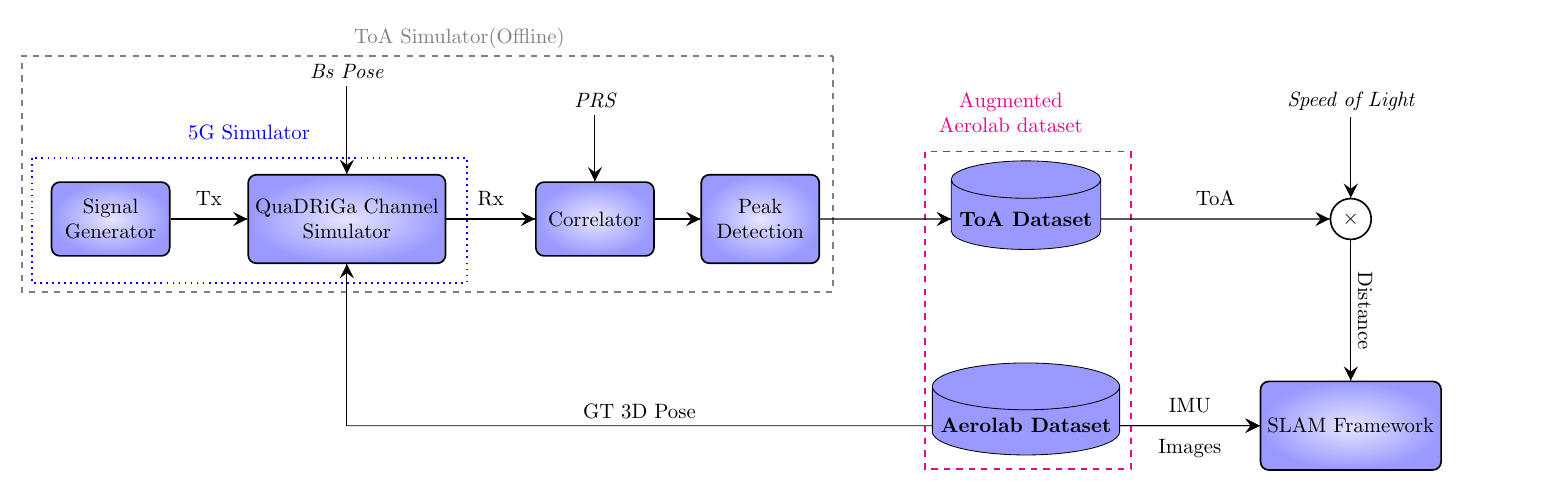}
    \caption[Dataset generation pipeline for 5G-enhanced SLAM]{Architecture of the dataset generation pipeline, showing the integration of simulated 5G ToA measurements with the Aerolab dataset. The combined data serves as input to the SLAM framework.}
    \label{fig:slam_architecture}
\end{figure*}

Due to infrastructure and hardware constraints, we employed a simulation-based approach to generate realistic ToA measurements. The simulation framework, illustrated in Fig.~\ref{fig:slam_architecture}, utilizes MATLAB 5G Toolbox for signal generation and QuaDRiGa \cite{jaeckel2017quadriga} for channel modeling, with two distinct 5G FR2 configurations:

\begin{itemize}
    \item 28 GHz: 200 MHz bandwidth, 120 kHz subcarrier spacing
    \item 78 GHz: 400 MHz bandwidth, 240 kHz subcarrier spacing
\end{itemize}

Four virtual base stations were positioned within the experimental space at coordinates BS1: $(2.5, -2.5, 4.5)$, BS2: $(2.5, 2.5, 4.0)$, BS3: $(-2.5, 2.5, 5.0)$, and BS4: $(-6.5, -2.5, 2.0)$ in the OptiTrack's global frame. Signals were generated at 10 Hz with consistent parameters (0 dBm transmit power, 10 dB SNR) and omnidirectional antennas.

\begin{table}[ht]
    \centering
    \caption[ToA stats for 28 GHz and 78 GHz across datasets and base stations]{ToA Mean and Standard Deviation for 28 GHz and 78 GHz Frequencies Across Five Datasets and Four Base Stations}
    \label{tab:toa_detailed_summary}
    \begin{adjustbox}{width=0.9\linewidth}
    \scriptsize
    \begin{tabular}{
        l        
        c        
        c        
        S[table-format=2.2] 
        S[table-format=2.2] 
    }
    \toprule
    \textbf{Dataset} & \textbf{Frequency (GHz)} & \textbf{Base Station} & \textbf{Std (cm)} & \textbf{Mean (cm)} \\
    \midrule
    \multirow{8}{*}{AL01} 
        & \multirow{4}{*}{28} 
            & \cellcolor{lightgray}1 & \cellcolor{lightgray}32.92 & \cellcolor{lightgray}4.20 \\
        &                       & 2 & 36.05 & -1.21 \\
        &                       & \cellcolor{lightgray}3 & \cellcolor{lightgray}31.99 & \cellcolor{lightgray}4.19 \\
        &                       & 4 & 41.32 & 8.28 \\
        \cmidrule(lr){2-5}
        & \multirow{4}{*}{78} 
            & \cellcolor{lightgray}1 & \cellcolor{lightgray}19.58 & \cellcolor{lightgray}-2.71 \\
        &                       & 2 & 19.44 & 1.34 \\
        &                       & \cellcolor{lightgray}3 & \cellcolor{lightgray}19.10 & \cellcolor{lightgray}0.31 \\
        &                       & 4 & 14.25 & -1.72 \\
    \midrule
    \multirow{8}{*}{AL02} 
        & \multirow{4}{*}{28} 
            & \cellcolor{lightgray}1 & \cellcolor{lightgray}36.31 & \cellcolor{lightgray}-6.88 \\
        &                       & 2 & 29.60 & -3.56 \\
        &                       & \cellcolor{lightgray}3 & \cellcolor{lightgray}36.60 & \cellcolor{lightgray}1.88 \\
        &                       & 4 & 32.22 & -2.68 \\
        \cmidrule(lr){2-5}
        & \multirow{4}{*}{78} 
            & \cellcolor{lightgray}1 & \cellcolor{lightgray}15.95 & \cellcolor{lightgray}-2.83 \\
        &                       & 2 & 17.01 & -2.29 \\
        &                       & \cellcolor{lightgray}3 & \cellcolor{lightgray}16.08 & \cellcolor{lightgray}0.56 \\
        &                       & 4 & 15.64 & -0.86 \\
    \midrule
    \multirow{8}{*}{AL03} 
        & \multirow{4}{*}{28} 
            & \cellcolor{lightgray}1 & \cellcolor{lightgray}38.44 & \cellcolor{lightgray}-8.24 \\
        &                       & 2 & 31.94 & -4.93 \\
        &                       & \cellcolor{lightgray}3 & \cellcolor{lightgray}33.25 & \cellcolor{lightgray}7.97 \\
        &                       & 4 & 30.49 & -0.50 \\
        \cmidrule(lr){2-5}
        & \multirow{4}{*}{78} 
            & \cellcolor{lightgray}1 & \cellcolor{lightgray}15.83 & \cellcolor{lightgray}0.87 \\
        &                       & 2 & 18.92 & -4.97 \\
        &                       & \cellcolor{lightgray}3 & \cellcolor{lightgray}16.38 & \cellcolor{lightgray}-0.80 \\
        &                       & 4 & 16.63 & -0.59 \\
    \midrule
    \multirow{8}{*}{AL04} 
        & \multirow{4}{*}{28} 
            & \cellcolor{lightgray}1 & \cellcolor{lightgray}40.14 & \cellcolor{lightgray}-14.56 \\
        &                       & 2 & 32.19 & -6.68 \\
        &                       & \cellcolor{lightgray}3 & \cellcolor{lightgray}31.17 & \cellcolor{lightgray}4.49 \\
        &                       & 4 & 27.64 & -4.97 \\
        \cmidrule(lr){2-5}
        & \multirow{4}{*}{78} 
            & \cellcolor{lightgray}1 & \cellcolor{lightgray}15.03 & \cellcolor{lightgray}0.87 \\
        &                       & 2 & 17.61 & -8.71 \\
        &                       & \cellcolor{lightgray}3 & \cellcolor{lightgray}14.69 & \cellcolor{lightgray}-1.49 \\
        &                       & 4 & 15.94 & -1.06 \\
    \midrule
    \multirow{8}{*}{AL05} 
        & \multirow{4}{*}{28} 
            & \cellcolor{lightgray}1 & \cellcolor{lightgray}38.23 & \cellcolor{lightgray}-19.41 \\
        &                       & 2 & 29.54 & -2.18 \\
        &                       & \cellcolor{lightgray}3 & \cellcolor{lightgray}31.23 & \cellcolor{lightgray}-3.09 \\
        &                       & 4 & 40.62 & -8.31 \\
        \cmidrule(lr){2-5}
        & \multirow{4}{*}{78} 
            & \cellcolor{lightgray}1 & \cellcolor{lightgray}14.40 & \cellcolor{lightgray}3.12 \\
        &                       & 2 & 16.80 & -4.67 \\
        &                       & \cellcolor{lightgray}3 & \cellcolor{lightgray}15.56 & \cellcolor{lightgray}0.57 \\
        &                       & 4 & 18.76 & -1.08 \\
    \bottomrule
    \end{tabular}%
    \end{adjustbox}
\end{table}

Table \ref{tab:toa_detailed_summary} shows the statistics of the simulated ToA measurements. The 78 GHz configuration demonstrates superior precision with standard deviations of 14.25-19.58 cm compared to 27.64-41.32 cm for 28 GHz. Mean errors at 78 GHz typically remain within ±5 cm, while 28 GHz exhibits larger biases up to 19.41 cm. This performance advantage can be attributed to the larger bandwidth and higher subcarrier spacing available at 78 GHz.


\subsection{Evaluation Metrics}

We evaluated our approach on both the Aerolab and EuRoC MAV datasets across multiple SLAM configurations (RGB-D, RGB-D-Inertial, Monocular, Monocular-Inertial, and Stereo) using the Absolute Trajectory Error (ATE) metric. For each configuration, we report:

\begin{itemize}
    \item \textbf{Local error (meters)}: Trajectory accuracy after SE3 post-alignment (rotation and translation).
    \item \textbf{Global error (meters)}: Absolute positioning accuracy without post-alignment.
\end{itemize}

For monocular configurations, which suffer from scale ambiguity, we additionally report:
\begin{itemize}
    \item \textbf{Local error}: Trajectory accuracy after Sim3 alignment (rotation, translation, and scale).
    \item \textbf{Unscaled local error}: Trajectory accuracy with only SE3 alignment (no scale correction).
    \item \textbf{Scale error (\%)}: Percentage deviation from true scale.
\end{itemize}

In local evaluation, the transformation between estimated trajectory and ground truth is calculated offline after SLAM completion, focusing on trajectory shape consistency. Global evaluation operates without post-processing alignment, requiring the system to maintain accurate absolute positioning throughout operation. To ensure statistical reliability, we conducted each experiment five times and reported the average ATE values across these runs.

\section{Results}
\label{sec:resuls}

\subsection{Global SLAM Performance with 5G \toa Integration}

This section presents the experimental results evaluating global SLAM performance with 5G \toa integration at two frequencies: 28 GHz and 78 GHz. We evaluate the system on two distinct datasets:

1. The Aerolab dataset: Performance metrics for RGB-D, RGB-D Inertial, and Monocular SLAM configurations (detailed in Table~\ref{tab:combined}).

2. The EuRoC MAV dataset: Stereo and Monocular SLAM performance across five sequences (V101-V103, V201-V202) (detailed in  Table~\ref{tab:euroc_combined}).
        

The tables include baseline measurements (without \toa integration) and results with \toa integration at both frequency bands, along with mode averages for each SLAM configuration.


\begin{table*}[htbp]
    \centering
   \caption[Comparison of SLAM Performance on Aerolab Dataset]{Global SLAM Performance with 5G ToA Integration on Aerolab Dataset. \\\textit{\footnotesize Note: For monocular mode with ToA, Datasets AL03 and AL05 results are calculated up to 75s and 65s, as tracking was lost beyond these points.}}
    \label{tab:combined}

    
    \begin{subtable}{0.45\linewidth}
        \centering
        \caption{RGB-D SLAM (SE3 Post-Alignment)}
       \resizebox{1\linewidth}{!}{%
    \begin{tabular}{c l c c c}
        \toprule
        \textbf{Dataset} & \textbf{Configuration}
          & \textbf{Local (m)} & \textbf{Global (m)} & \textbf{Local Imp. } \\
        \midrule

        \multirow{3}{*}{AL01}
            & Baseline
                & 0.064 & N/A{$^\ddagger$}    & -- \\
            & \cellcolor{lightgray}28 GHz \toa
                & \cellcolor{lightgray}0.066
                & \cellcolor{lightgray}0.237
                & \cellcolor{lightgray}\coloredpercent{-3.1} \\
            & 78 GHz \toa
                & 0.055 & 0.101 & \coloredpercent{14.1} \\
        \cmidrule(lr){1-5}

        \multirow{3}{*}{AL02}
            & Baseline
                & 0.116 & N/A    & -- \\
            & \cellcolor{lightgray}28 GHz \toa
                & \cellcolor{lightgray}0.117
                & \cellcolor{lightgray}0.165
                & \cellcolor{lightgray}\coloredpercent{-0.9} \\
            & 78 GHz \toa
                & 0.117 & 0.133 & \coloredpercent{-0.9} \\
        \cmidrule(lr){1-5}

        \multirow{3}{*}{AL03}
            & Baseline
                & 0.155 & N/A    & -- \\
            & \cellcolor{lightgray}28 GHz \toa
                & \cellcolor{lightgray}0.160
                & \cellcolor{lightgray}0.197
                & \cellcolor{lightgray}\coloredpercent{-3.2} \\
            & 78 GHz \toa
                & 0.148 & 0.158 & \coloredpercent{4.5} \\
        \cmidrule(lr){1-5}

        \multirow{3}{*}{AL04}
            & Baseline
                & 0.095 & N/A    & -- \\
            & \cellcolor{lightgray}28 GHz \toa
                & \cellcolor{lightgray}0.102
                & \cellcolor{lightgray}0.149
                & \cellcolor{lightgray}\coloredpercent{-7.4} \\
            & 78 GHz \toa
                & 0.094 & 0.104 & \coloredpercent{1.1} \\
        \cmidrule(lr){1-5}

        \multirow{3}{*}{AL05}
            & Baseline
                & 0.146 & N/A    & -- \\
            & \cellcolor{lightgray}28 GHz \toa
                & \cellcolor{lightgray}0.162
                & \cellcolor{lightgray}0.242
                & \cellcolor{lightgray}\coloredpercent{-11.0} \\
            & 78 GHz \toa
                & 0.164 & 0.171 & \coloredpercent{-12.3} \\
        \bottomrule

        \multirow{3}{*}{\textbf{Mode Average}}
            & Baseline
                & \textbf{0.115} & N/A & -- \\
            & \cellcolor{lightgray}28 GHz \toa
                & \cellcolor{lightgray}\textbf{0.121}
                & \cellcolor{lightgray}\textbf{0.198}
                & \cellcolor{lightgray}\textbf{\coloredpercent{-5.2}} \\
            & 78 GHz \toa
                & \textbf{0.116} & \textbf{0.133} & \textbf{\coloredpercent{1.3}} \\

        \bottomrule
    \end{tabular}
        }
    \end{subtable}
    \hspace{0mm}
    \begin{subtable}{0.45\linewidth}
        \centering
        \caption{RGB-D Inertial SLAM (SE3 Post-Alignment)}
       \resizebox{1\linewidth}{!}{%
    \begin{tabular}{c l c c c}
        \toprule
        \textbf{Dataset} & \textbf{Configuration}
          & \textbf{Local (m)} & \textbf{Global (m)} & \textbf{Local Imp. } \\
        \midrule
    \multirow{3}{*}{AL01} 
    & Baseline
        & 0.089 & N/A    & -- \\
    & \cellcolor{lightgray}28 GHz \toa
        & \cellcolor{lightgray}0.092
        & \cellcolor{lightgray}0.471
        & \cellcolor{lightgray}\coloredpercent{-3.4} \\
    & 78 GHz \toa
        & 0.075 & 0.276 & \coloredpercent{15.7} \\
\cmidrule(lr){1-5}

\multirow{3}{*}{AL02} 
    & Baseline
        & 0.119 & N/A    & -- \\
    & \cellcolor{lightgray}28 GHz \toa
        & \cellcolor{lightgray}0.118
        & \cellcolor{lightgray}0.206
        & \cellcolor{lightgray}\coloredpercent{0.8} \\
    & 78 GHz \toa
        & 0.118 & 0.315 & \coloredpercent{0.8} \\
\cmidrule(lr){1-5}

\multirow{3}{*}{AL03} 
    & Baseline
        & 0.216 & N/A    & -- \\
    & \cellcolor{lightgray}28 GHz \toa
        & \cellcolor{lightgray}0.231$^\dagger$
        & \cellcolor{lightgray}0.267
        & \cellcolor{lightgray}\coloredpercent{-6.9} $^\dagger$ \\
    & 78 GHz \toa
        & 0.366$^\dagger$ & 0.426 & \coloredpercent{-69.4} $^\dagger$ \\
\cmidrule(lr){1-5}

\multirow{3}{*}{AL04} 
    & Baseline
        & 0.065 & N/A    & -- \\
    & \cellcolor{lightgray}28 GHz \toa
        & \cellcolor{lightgray}0.064
        & \cellcolor{lightgray}0.134
        & \cellcolor{lightgray}\coloredpercent{1.5} \\
    & 78 GHz \toa
        & 0.062 & 0.125 & \coloredpercent{4.6} \\
\cmidrule(lr){1-5}

\multirow{3}{*}{AL05} 
    & Baseline
        & 0.108 & N/A    & -- \\
    & \cellcolor{lightgray}28 GHz \toa
        & \cellcolor{lightgray}0.105
        & \cellcolor{lightgray}0.321
        & \cellcolor{lightgray}\coloredpercent{2.8} \\
    & 78 GHz \toa
        & 0.105 & 0.168 & \coloredpercent{2.8} \\
\bottomrule

\multirow{3}{*}{\textbf{Mode Average}} 
    & Baseline
        & \textbf{0.095} & N/A & -- \\
    & \cellcolor{lightgray}28 GHz \toa
        & \cellcolor{lightgray}\textbf{0.94}
        & \cellcolor{lightgray}\textbf{0.280}
        & \cellcolor{lightgray}\textbf{\coloredpercent{0.42}} \\
    & 78 GHz \toa
        & \textbf{0.90} & \textbf{0.262} & \textbf{\coloredpercent{5.97}} \\
        \bottomrule

        \end{tabular}
        }
    \end{subtable}

    \vspace{0.4cm}

    \begin{subtable}{0.65\linewidth}
        \centering
        \caption{Monocular SLAM (Both SE3 and Sim3 Post-Alignment for \toa)}
       \resizebox{1\linewidth}{!}{%

       \begin{tabular}{c l c c c c c}
        \toprule
        \textbf{Dataset} & \textbf{Configuration}
          & \textbf{Local (m)} & \textbf{Global (m)} & \textbf{Local (Unscaled)} & \textbf{Scale} & \textbf{Local Imp.} \\
        \midrule
    
    \multirow{3}{*}{AL01} 
        & Baseline 
            & Failed    & N/A & N/A & N/A & -- \\
        & \cellcolor{lightgray}28 GHz \toa
            & \cellcolor{lightgray}Failed 
            & \cellcolor{lightgray}Failed 
            & \cellcolor{lightgray}-- 
            & \cellcolor{lightgray}-- 
            & \cellcolor{lightgray}-- \\
        & 78 GHz \toa
            & Failed    & Failed & -- & -- & -- \\
    \cmidrule(lr){1-7}
    
    \multirow{3}{*}{AL02} 
        & Baseline
            & 0.055 & N/A    & N/A & N/A & -- \\
        & \cellcolor{lightgray}28 GHz \toa
            & \cellcolor{lightgray}0.055
            & \cellcolor{lightgray}0.380
            & \cellcolor{lightgray}0.059 
            & \cellcolor{lightgray}0.93\% 
            & \cellcolor{lightgray}\coloredpercent{0.6} \\
        & 78 GHz \toa
            & 0.057 & 0.121 & 0.058 & 0.37\% & \coloredpercent{-3.3} \\
    \cmidrule(lr){1-7}
    
    \multirow{3}{*}{AL03} 
        & Baseline
            & 0.058 & N/A    & N/A & N/A & -- \\
        & \cellcolor{lightgray}28 GHz \toa
            & \cellcolor{lightgray}0.084
            & \cellcolor{lightgray}0.186
            & \cellcolor{lightgray}0.089 
            & \cellcolor{lightgray}1.58\% 
            & \cellcolor{lightgray}\coloredpercent{-44.0}$^\dagger$ \\
        & 78 GHz \toa
            & 0.074 & 0.152 & 0.075 & 0.62\% & \coloredpercent{-26.6}$^\dagger$ \\
    \cmidrule(lr){1-7}
    
    \multirow{3}{*}{AL04} 
        & Baseline
            & 0.074 & N/A    & N/A & N/A & -- \\
        & \cellcolor{lightgray}28 GHz \toa
            & \cellcolor{lightgray}0.077
            & \cellcolor{lightgray}0.162
            & \cellcolor{lightgray}0.084 
            & \cellcolor{lightgray}1.97\% 
            & \cellcolor{lightgray}\coloredpercent{-3.8} \\
        & 78 GHz \toa
            & 0.056 & 0.127 & 0.058 & 0.76\% & \coloredpercent{23.8} \\
    \cmidrule(lr){1-7}
    
    \multirow{3}{*}{AL05} 
        & Baseline
            & 0.062 & N/A    & N/A & N/A & -- \\
        & \cellcolor{lightgray}28 GHz \toa
            & \cellcolor{lightgray}0.075
            & \cellcolor{lightgray}0.260
            & \cellcolor{lightgray}0.084 
            & \cellcolor{lightgray}2.44\% 
            & \cellcolor{lightgray}\coloredpercent{-20.2} \\
        & 78 GHz \toa
            & 0.054 & 0.131 & 0.061 & 1.90\% & \coloredpercent{13.8} \\
    \cmidrule(lr){1-7}
    
    \multirow{3}{*}{\textbf{Mode Average}} 
        & Baseline
            & \textbf{0.064} & N/A & N/A & N/A & -- \\
        & \cellcolor{lightgray}28 GHz \toa
            & \cellcolor{lightgray}\textbf{0.069}
            & \cellcolor{lightgray}\textbf{0.267}
            & \cellcolor{lightgray}\textbf{0.076} 
            & \cellcolor{lightgray}\textbf{0.30\%} 
            & \cellcolor{lightgray}\textbf{\coloredpercent{-7.8}} \\
        & 78 GHz \toa
            & \textbf{0.056} & \textbf{0.126} & \textbf{0.059} & \textbf{0.15\%} & \textbf{\coloredpercent{11.4}} \\
    \bottomrule
    \end{tabular}

        }
    \end{subtable}

        \vspace{0.3em}
\begin{minipage}{0.65\linewidth}
        {\fontsize{6}{8} \selectfont
            $\dagger$ AL03's RGB-D Inertial and monocular \toa results were excluded from average  due to anomalous degradation. \\[-0.4em]
            $\ddagger$ N/A denotes baseline systems without global positioning and scaling capability. \\[-0.4em]
            Monocular-inertial configurations failed across all datasets due to initialization issues, preventing successful tracking.\\[-0.4em]
            For RGB-D and RGB-D Inertial modes, accuracy is evaluated using SE3 post-alignment. \\ [-0.4em]
            For Monocular mode, the baseline uses Sim3 alignment, while ToA configurations provide both SE3 and Sim3 alignment results.}
    \end{minipage}
\end{table*}

\begin{table}[htbp]
    \centering
    \caption[Comparison of SLAM Performance on EuRoC MAV Dataset]{Average Errors and Improvements for Different SLAM Modes on EuRoC MAV Dataset. \textit{\footnotesize \\\textit{\footnotesize Note: Results for V203  are excluded  due to tracking failures.}}}
    \label{tab:euroc_combined}

\begin{subtable}{0.9\linewidth}
    \centering
    \caption{Stereo SLAM (SE3 Post-Alignment for \toa)}
    \resizebox{1\linewidth}{!}{%
    \begin{tabular}{c l c c c}
        \toprule
        \textbf{Sequence} & \textbf{Configuration}
          & \textbf{Local (m)} & \textbf{Global (m)} & \textbf{Local Imp.} \\
        \midrule

        \multirow{3}{*}{V101} 
            & Baseline & 0.084 & N/A & -- \\
            & \cellcolor{lightgray}28 GHz \toa
                & \cellcolor{lightgray}0.084
                & \cellcolor{lightgray}0.232
                & \cellcolor{lightgray}\coloredpercent{0.0} \\
            & 78 GHz \toa
                & 0.081 & 0.183 & \coloredpercent{3.6} \\
        \cmidrule(lr){1-5}

        \multirow{3}{*}{V102} 
            & Baseline & 0.099 & N/A & -- \\
            & \cellcolor{lightgray}28 GHz \toa
                & \cellcolor{lightgray}0.113
                & \cellcolor{lightgray}0.585
                & \cellcolor{lightgray}\coloredpercent{-14.1} \\
            & 78 GHz \toa
                & 0.106 & 0.139 & \coloredpercent{-7.2} \\
        \cmidrule(lr){1-5}

        \multirow{3}{*}{V103} 
            & Baseline & 0.161 & N/A & -- \\
            & \cellcolor{lightgray}28 GHz \toa
                & \cellcolor{lightgray}0.182
                & \cellcolor{lightgray}0.359
                & \cellcolor{lightgray}\coloredpercent{-13.0} \\
            & 78 GHz \toa
                & 0.126 & 0.190 & \coloredpercent{21.7} \\
        \cmidrule(lr){1-5}

        \multirow{3}{*}{V201} 
            & Baseline & 0.113 & N/A & -- \\
            & \cellcolor{lightgray}28 GHz \toa
                & \cellcolor{lightgray}0.094
                & \cellcolor{lightgray}0.335
                & \cellcolor{lightgray}\coloredpercent{17.2} \\
            & 78 GHz \toa
                & 0.086 & 0.177 & \coloredpercent{23.8} \\
        \cmidrule(lr){1-5}

        \multirow{3}{*}{V202} 
            & Baseline & 0.190 & N/A & -- \\
            & \cellcolor{lightgray}28 GHz \toa
                & \cellcolor{lightgray}0.196
                & \cellcolor{lightgray}0.274
                & \cellcolor{lightgray}\coloredpercent{-3.0} \\
            & 78 GHz \toa
                & 0.192 & 0.265 & \coloredpercent{-1.0} \\
        \cmidrule(lr){1-5}

        \multirow{3}{*}{\textbf{Average}$^*$} 
            & Baseline & \textbf{0.129} & N/A & -- \\
            & \cellcolor{lightgray}28 GHz \toa
                & \cellcolor{lightgray}\textbf{0.134}
                & \cellcolor{lightgray}\textbf{0.357}
                & \cellcolor{lightgray}\textbf{\coloredpercent{-2.6}} \\
            & 78 GHz \toa
                & \textbf{0.118} & \textbf{0.191} & \textbf{\coloredpercent{8.2}} \\
        \bottomrule
    \end{tabular}
    }
\end{subtable}
    \vfill
    \vspace{3mm}
    \begin{subtable}{1\linewidth}
        \centering
        \caption{Monocular SLAM (SE3 and Sim3 Post-Alignment for \toa)}
        \resizebox{1\linewidth}{!}{%

       \begin{tabular}{c l c c c c c}
    \toprule
    \textbf{Sequence} & \textbf{Configuration}
      & \textbf{Local (m)} & \textbf{Global (m)} & \textbf{Local (Unscaled) (m)} & \textbf{Scale} & \textbf{Local Imp.} \\
    \midrule
\multirow{3}{*}{V101}
    & Baseline & 0.095 & N/A & N/A & N/A & -- \\
    & \cellcolor{lightgray}28 GHz \toa
        & \cellcolor{lightgray}0.094
        & \cellcolor{lightgray}0.247
        & \cellcolor{lightgray}0.099
        & \cellcolor{lightgray}1.55\%
        & \cellcolor{lightgray}\coloredpercent{1.1} \\
    & 78 GHz \toa
        & 0.094 & 0.149 & 0.094 & 0.05\% & \coloredpercent{1.1} \\
\cmidrule(lr){1-7}

\multirow{3}{*}{V102} 
    & Baseline & 0.100 & N/A & N/A & N/A & -- \\
    & \cellcolor{lightgray}28 GHz \toa
        & \cellcolor{lightgray}0.100
        & \cellcolor{lightgray}0.438
        & \cellcolor{lightgray}0.102
        & \cellcolor{lightgray}0.40\%
        & \cellcolor{lightgray}\coloredpercent{0.0} \\
    & 78 GHz \toa
        & 0.098 & 0.128 & 0.098 & 0.59\% & \coloredpercent{2.0} \\
\cmidrule(lr){1-7}

\multirow{3}{*}{V103} 
    & Baseline & 0.116 & N/A & N/A & N/A & -- \\
    & \cellcolor{lightgray}28 GHz \toa
        & \cellcolor{lightgray}0.103
        & \cellcolor{lightgray}0.224
        & \cellcolor{lightgray}0.111
        & \cellcolor{lightgray}2.24\%
        & \cellcolor{lightgray}\coloredpercent{11.2} \\
    & 78 GHz \toa
        & 0.102 & 0.415 & 0.102 & 0.01\% & \coloredpercent{12.1} \\
\cmidrule(lr){1-7}

\multirow{3}{*}{V201} 
    & Baseline & 0.086 & N/A & N/A & N/A & -- \\
    & \cellcolor{lightgray}28 GHz \toa
        & \cellcolor{lightgray}0.085
        & \cellcolor{lightgray}0.669
        & \cellcolor{lightgray}0.125
        & \cellcolor{lightgray}3.75\%
        & \cellcolor{lightgray}\coloredpercent{1.2} \\
    & 78 GHz \toa
        & 0.085 & 0.247 & 0.085 & 0.17\% & \coloredpercent{1.2} \\
\cmidrule(lr){1-7}

\multirow{3}{*}{V202} 
    & Baseline & 0.176 & N/A & N/A & N/A & -- \\
    & \cellcolor{lightgray}28 GHz \toa
        & \cellcolor{lightgray}0.178
        & \cellcolor{lightgray}0.579
        & \cellcolor{lightgray}0.179
        & \cellcolor{lightgray}0.74\%
        & \cellcolor{lightgray}\coloredpercent{-1.1} \\
    & 78 GHz \toa
        & 0.178 & 0.248 & 0.181 & 1.36\% & \coloredpercent{-1.1} \\
\cmidrule(lr){1-7}

\multirow{3}{*}{\textbf{Average}$^*$} 
    & Baseline & \textbf{0.115} & N/A & N/A & N/A & -- \\
    & \cellcolor{lightgray}28 GHz \toa
        & \cellcolor{lightgray}\textbf{0.112}
        & \cellcolor{lightgray}\textbf{0.431}
        & \cellcolor{lightgray}\textbf{0.123}
        & \cellcolor{lightgray}\textbf{1.74\%}
        & \cellcolor{lightgray}\textbf{\coloredpercent{2.5}} \\
    & 78 GHz \toa
        & \textbf{0.111} & \textbf{0.237} & \textbf{0.112} & \textbf{0.20\%} & \textbf{\coloredpercent{3.1}} \\
\bottomrule
\end{tabular}

        }
    \end{subtable}
\end{table}




Before delving into the analysis, it is crucial to understand the difference between local and global evaluations and the implications of including baseline global results. In the \textbf{local evaluation}, the transformation (rotation and translation) between the estimated trajectory and the ground truth is assumed to be unknown and is calculated offline after the SLAM process. This alignment is typically performed using methods like Horn's absolute orientation algorithm, which finds the best-fit transformation that minimizes the trajectory error. Essentially, the local evaluation focuses on the consistency of the trajectory shape, disregarding the absolute positioning and orientation in the global frame.

\begin{figure}[!htbp]
    \centering
    \includegraphics[width=0.5\textwidth]{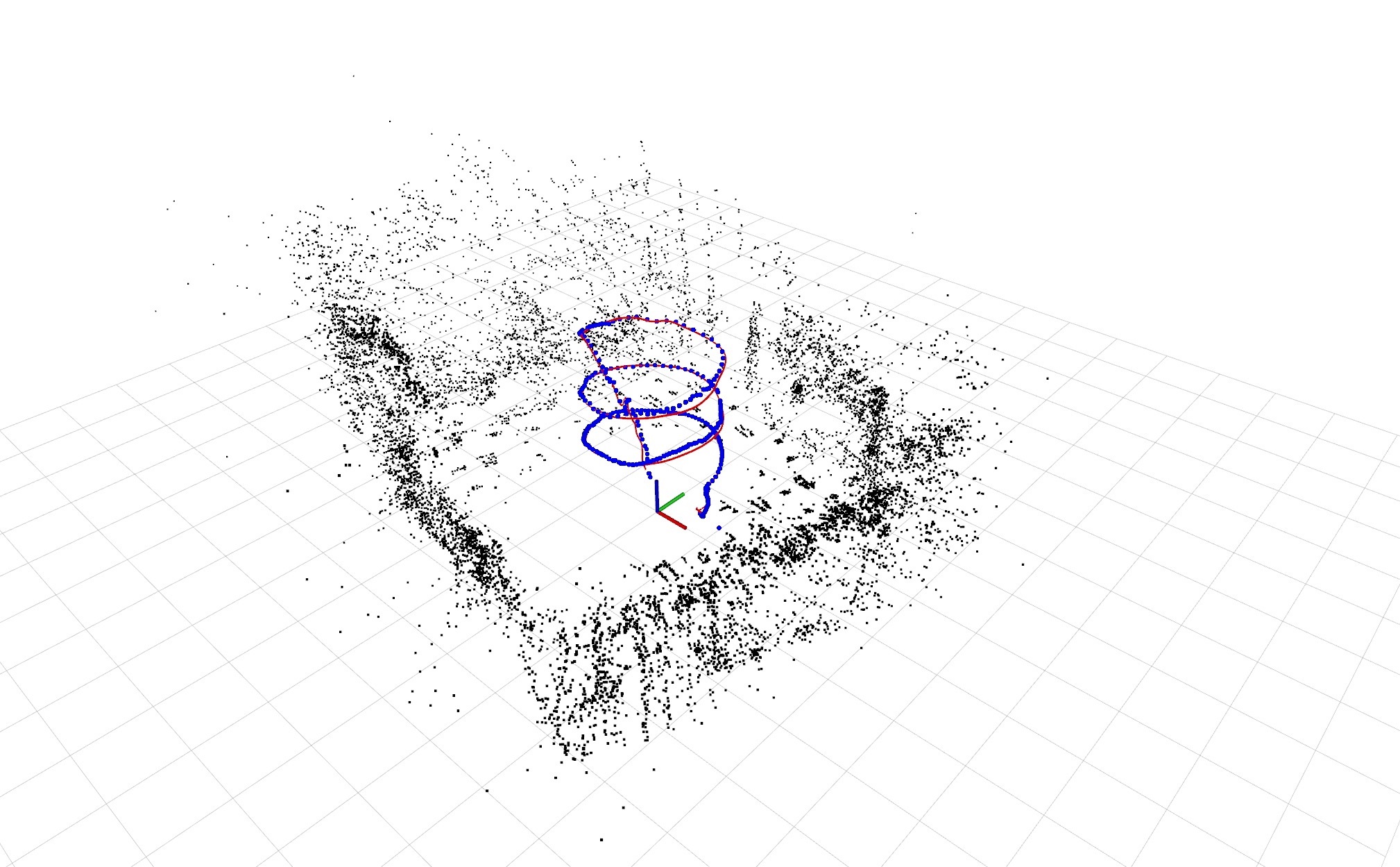}
    \caption[Global SLAM estimate with \toa integration]{RGB-D trajectory and map point estimates in the global frame for AL01 with 78 GHz \toa integration.
        The SLAM results show keyframe pose estimates (blue lines), mapped features (black dots), and the ground truth trajectory (red line). Coordinate axes are color-coded: red for X, green for Y, and blue for Z. The trajectory estimate, enhanced with 78 GHz \toa data, aligns closely with the ground truth in the global frame, demonstrating the system's ability to maintain accurate global positioning without post-processing alignment.}
    \label{fig:3d_global}
\end{figure}

In contrast, the \textbf{global evaluation} does not perform any post-processing alignment. The SLAM algorithm must estimate both the trajectory and the transformation directly, operating entirely in the global coordinate frame. This approach is more challenging because it requires the SLAM system to maintain accurate global positioning and orientation throughout the operation without relying on offline corrections. This capability is demonstrated in the 3D plot shown in Fig.~\ref{fig:3d_global}, which showcases one experiment (Dataset 0). The RGB-D trajectory enhanced with 78 GHz \toa measurements in the global frame closely matches the ground truth, demonstrating how \toa integration enables consistent global positioning.

\subsubsection{Analysis of SLAM Modes with \toa Integration}
The integration of \toa measurements into various SLAM modes reveals several key insights across both the Aerolab and EuRoC datasets. For RGB-D SLAM, the baseline already achieves good local accuracy (0.115 m average), while \toa integration shows inconsistent effects—78 GHz \toa yields a modest improvement of 1.3\% in local accuracy, whereas 28 GHz \toa causes a 5.2\% degradation. AL01 exhibits the most significant improvement (14.1\%) with 78 GHz \toa, contrasting with AL05's 12.3\% degradation using the same configuration.

RGB-D Inertial SLAM follows a similar pattern with more pronounced variations: 78 GHz \toa shows an average local improvement of 5.97\%, while 28 GHz \toa offers minimal improvement (0.42\%). AL03 presents anomalous degradation with both \toa frequencies (-6.9\% and -69.4\%, excluded from averages), indicating potential trajectory-specific challenges.

In Monocular SLAM, where scale estimation is inherently problematic, 78 GHz \toa integration provides a substantial 11.4\% improvement in local accuracy while maintaining excellent scale estimation (0.156\% average error). Meanwhile, 28 GHz \toa integration results in a 7.8\% degradation despite reasonable scale estimation (0.300\% error). On the EuRoC dataset, both Stereo (8.2\% improvement with 78 GHz) and Monocular modes (3.1\% improvement with 78 GHz) demonstrate consistent benefits from 78 GHz \toa integration.

The most significant advantage of \toa integration lies in enabling global positioning across all SLAM modes—a capability absent in baseline configurations. The 78 GHz \toa consistently outperforms 28 GHz \toa in global accuracy: RGB-D (0.133 m vs. 0.198 m), RGB-D Inertial (0.262 m vs. 0.280 m), and Monocular (0.126 m vs. 0.267 m) on the Aerolab dataset. This performance disparity  stems from the higher precision of 78 GHz measurements.

Environmental factors and trajectory characteristics significantly influence \toa integration performance, as evidenced by the inconsistent results across datasets. Dataset 2 particularly demonstrates how specific conditions can challenge \toa integration across multiple SLAM modes. Despite these variations, the addition of global positioning capabilities represents a substantial enhancement for applications requiring absolute positioning, even when local accuracy improvements remain marginal or inconsistent.

\subsection{SLAM Performance with Unknown Base Station Configurations}

To evaluate SLAM performance with unknown base station locations, we modified our system architecture by changing the base station nodes from fixed to non-fixed status while maintaining a fixed transformation node. This configuration eliminates the requirement for prior knowledge of base station locations, though it does preclude global positioning capabilities. We further relaxed operational conditions by:
\begin{itemize}
\item Reducing the number of active base stations
\item Implementing sequential line-of-sight (LOS) connections where only one base station maintains LOS during specific, non-overlapping time intervals
\item Limiting base station coverage to partial trajectory segments rather than the complete path
\end{itemize}
We evaluated these scenarios across both minimal and rich sensor configurations under 78 GHz 5G network. For minimal sensor setups, we focused on monocular configurations using both the Aerolab and EuRoC MAV datasets, as \toa integration typically shows the most pronounced effects with limited sensor data. For rich sensor configurations (RGB-D and stereo), we investigated \toa's potential as an alternative to loop closure.
Loop closure may be unavailable in many practical scenarios, such as:
\begin{itemize}
\item Linear trajectories without path intersections
\item Exploration of previously unvisited areas
\item Time-critical operations that prevent revisiting locations
\item Environments with significant appearance changes
\item Resource-constrained systems where loop closure detection is computationally prohibitive
\end{itemize}

We identified sequences where loop closure was naturally occurring and contributing to the system's accuracy (AL01, V102, V103, V202, and V203). To evaluate \toa measurements as a potential alternative to loop closure for drift correction, we deliberately disabled loop closure in these specific sequences. This approach allowed us to directly assess whether ToA integration could effectively compensate for the absence of loop closure in practical scenarios where visual loop closure might be unavailable or unreliable.

For comparison with recent literature, we also replicated the experimental setup from \cite{li2023uwb} using the EuRoC Machine Hall dataset. While their work used simulated \toa measurements from one base station at 30 Hz ($\sigma = 5$ cm Gaussian noise, base station at [10,10,10]), we modified the scenario to operate at a more practical 5 Hz measurement frequency. This allows direct comparison while reflecting more realistic operational constraints.

\subsubsection{Sequential Unknown Base Station Operation (Monocular)}
\begin{table}[h]
    \centering
    \caption{MONO SLAM with Sequential Unknown Base Stations}
    \label{tab:comprehensive_comparison}
    \begin{adjustbox}{width=1\linewidth}
    \begin{tabular}{c l c c c c}
        \toprule
        \textbf{Sequence} & \textbf{Configuration}
          & \textbf{Local (m)} & \textbf{Local (Unscaled) (m)} & \textbf{Scale} & \textbf{Local Imp.} \\
        \midrule

        \multirow{2}{*}{V101} 
            & Baseline & 0.0950 & N/A & N/A & -- \\
            & \cellcolor{lightgray}Sequential BS
                & \cellcolor{lightgray}0.0944
                & \cellcolor{lightgray}0.1222
                & \cellcolor{lightgray}3.81\%
                & \cellcolor{lightgray}\coloredpercent{0.63} \\
        \cmidrule(lr){1-6}

        \multirow{2}{*}{V102} 
            & Baseline & 0.1004 & N/A & N/A & -- \\
            & \cellcolor{lightgray}Sequential BS
                & \cellcolor{lightgray}0.1001
                & \cellcolor{lightgray}0.1020
                & \cellcolor{lightgray}0.71\%
                & \cellcolor{lightgray}\coloredpercent{0.30} \\
        \cmidrule(lr){1-6}

        \multirow{2}{*}{V103} 
            & Baseline & 0.1160 & N/A & N/A & -- \\
            & \cellcolor{lightgray}Sequential BS
                & \cellcolor{lightgray}0.1068
                & \cellcolor{lightgray}0.1144
                & \cellcolor{lightgray}1.63\%
                & \cellcolor{lightgray}\coloredpercent{7.93} \\
        \cmidrule(lr){1-6}

        \multirow{2}{*}{V201} 
            & Baseline & 0.0860 & N/A & N/A & -- \\
            & \cellcolor{lightgray}Sequential BS
                & \cellcolor{lightgray}0.0848
                & \cellcolor{lightgray}0.0924
                & \cellcolor{lightgray}1.33\%
                & \cellcolor{lightgray}\coloredpercent{1.39} \\
        \cmidrule(lr){1-6}

        \multirow{2}{*}{V202} 
            & Baseline & 0.1760 & N/A & N/A & -- \\
            & \cellcolor{lightgray}Sequential BS
                & \cellcolor{lightgray}0.1748
                & \cellcolor{lightgray}0.1803
                & \cellcolor{lightgray}0.38\%
                & \cellcolor{lightgray}\coloredpercent{0.68} \\
        \cmidrule(lr){1-6}

        \multirow{2}{*}{AL02} 
            & Baseline & 0.0550 & N/A & N/A & -- \\
            & \cellcolor{lightgray}Sequential BS
                & \cellcolor{lightgray}0.0534
                & \cellcolor{lightgray}0.0709
                & \cellcolor{lightgray}0.90\%
                & \cellcolor{lightgray}\coloredpercent{2.90} \\
        \cmidrule(lr){1-6}

        \multirow{2}{*}{AL04} 
            & Baseline & 0.0740 & N/A & N/A & -- \\
            & \cellcolor{lightgray}Sequential BS
                & \cellcolor{lightgray}0.0612
                & \cellcolor{lightgray}0.0714
                & \cellcolor{lightgray}0.31\%
                & \cellcolor{lightgray}\coloredpercent{17.29} \\
        \cmidrule(lr){1-6}

        \multirow{2}{*}{\textbf{Average}} 
            & Baseline & \textbf{0.1003} & N/A & N/A & -- \\
            & \cellcolor{lightgray}Sequential BS
                & \cellcolor{lightgray}\textbf{0.0965}
                & \cellcolor{lightgray}\textbf{0.1077}
                & \cellcolor{lightgray}\textbf{1.30\%}
                & \cellcolor{lightgray}\textbf{\coloredpercent{4.40}} \\
        \bottomrule
    \end{tabular}
     \end{adjustbox}
    \vspace{0.3em}
    \begin{minipage}{1\linewidth}
        {\fontsize{6}{9} \selectfont
            ~ \\[-0.4em]
            Sequential BS: Three unknown base stations operating in different time intervals  \\[-0.4em]
            Local: Post-scale alignment error; Local (Unscaled): Pre-scale alignment error. \\[-0.4em]
            Improvement was calculated relative to baseline performance. \\[-0.4em]
            Note: V203, AL01, AL03, and AL05 lose tracking before 100 seconds.
        }
    \end{minipage}
\end{table}

\begin{figure}[!htbp]
    \centering
    \includegraphics[width=1\linewidth]{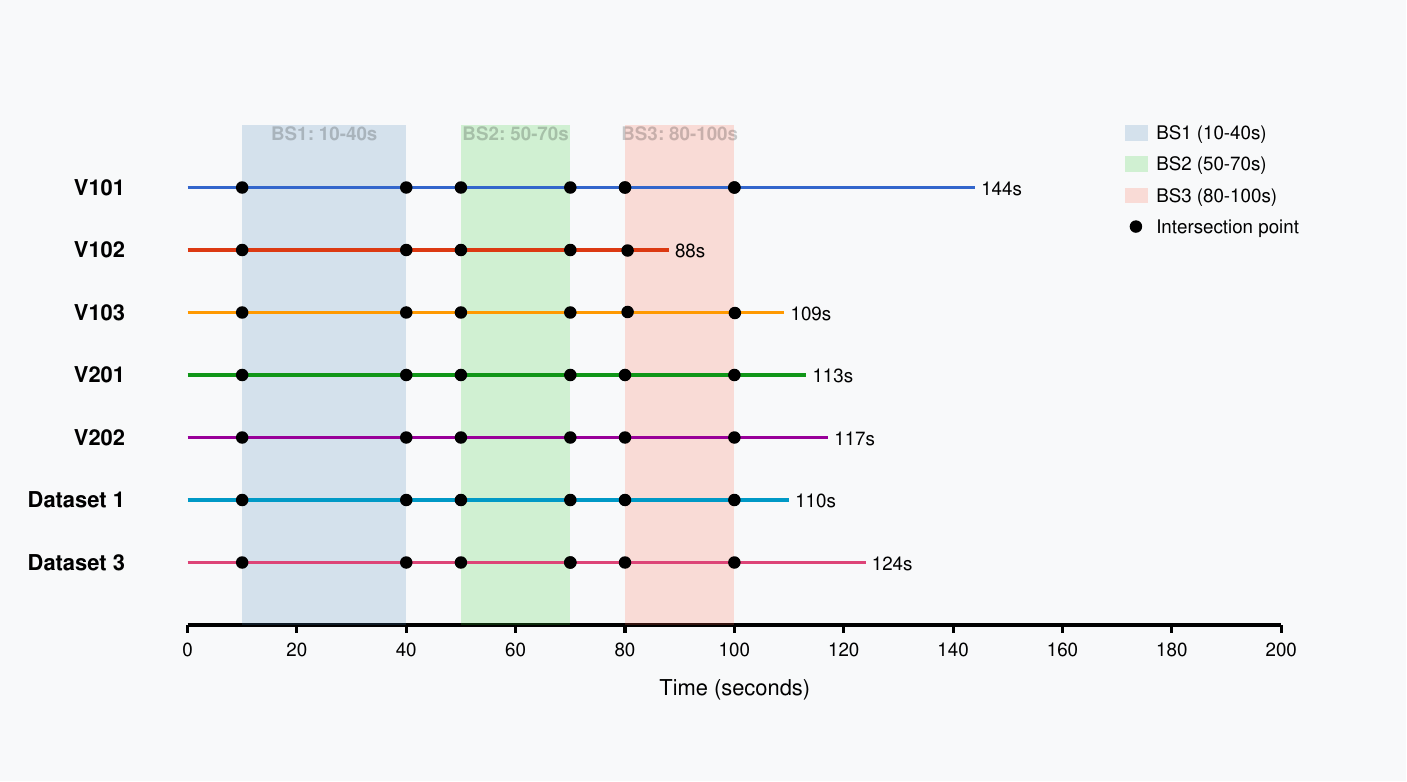}
    \caption[Bastion Station Activity Over Trajectories (Euroc MAV and Aerolab)]{Bastion Station Activity Over Trajectories: 
    This plot illustrates the active periods of three bastion stations (BS1: 10–40s, BS2: 50–70s, BS3: 80–100s)
     over the trajectories of various datasets (V101, V102, V103, V201, V202, Dataset 1, and Dataset 3). 
     Each horizontal bar represents the total duration of a dataset, with colored segments highlighting the intervals
      during which each bastion station is active}
      \label{fig:mono_seq_timeline}
\end{figure}

Table \ref{tab:comprehensive_comparison} presents results for scenarios where three base stations operate in different time intervals (BS1: 10-40s, BS2: 50-70s, BS3: 80-100s) with unknown locations at frequency 78 GHz as illustrated in Fig.~\ref{fig:mono_seq_timeline}. For the monocular configuration, results demonstrate consistent improvement across datasets despite the challenging sequential operation scenario. We observed an average improvement of 4.40\% in local accuracy across all sequences, with particularly strong improvements in AL04 (17.29\%) and V103 (7.93\%). The system maintained reliable scale estimation with an average error of 1.30\% across all datasets. These improvements are especially noteworthy considering the lack of continuous base station coverage, suggesting that even intermittent \toa measurements can significantly enhance SLAM performance in minimal sensor configurations.

\subsubsection{\toa Integration as an Alternative to Loop Closure}
In typical SLAM implementations, loop closure serves as a critical mechanism for drift correction by recognizing previously visited locations. However, this capability faces limitations in many real-world scenarios. Our investigation focuses on whether \toa measurements can provide a viable alternative when traditional loop closure is unavailable or unreliable. This study is conducted using RGBD for the Aerolab dataset and Stereo for the Euroc Mav dataset.

Table \ref{tab:toa_effect_local} presents a comprehensive comparison between baseline SLAM with loop closure enabled, the same system with loop closure disabled, and two \toa integration configurations using unknown base station locations: sequential operation of three base stations and continuous operation of three base stations. The results reveal a compelling case for \toa integration as a loop closure alternative.

\begin{table*}[htbp]
    \centering
    \caption{ToA Integration as an Alternative to Loop Closure}
    \label{tab:toa_effect_local}
    \begin{adjustbox}{width=0.7\linewidth}
    \begin{tabular}{l c c c c c}
        \toprule
        \textbf{Dataset} & \textbf{Baseline (LC)} & \textbf{Baseline (No LC)} & \textbf{\toa (Seq)} & \textbf{\toa (Cont)} & \textbf{Improvement (Seq / Cont) (\%)} \\
        \midrule

        AL01 
            & 0.064 & 0.153 & 0.138 & 0.129 & \coloredpercent{9.8} / \coloredpercent{15.7} \\

        \rowcolor{lightgray}
        V102 
            & 0.099 & 0.272 & 0.205 & 0.142 & \coloredpercent{24.6} / \coloredpercent{47.8} \\

        V103 
            & 0.161 & 0.507 & 0.299 & 0.257 & \coloredpercent{41.0} / \coloredpercent{49.3} \\

        \rowcolor{lightgray}
        V202 
            & 0.190 & 0.213 & 0.230 & 0.208 & \coloredpercent{-8.0} / \coloredpercent{2.3} \\

        V203 
            & 0.960 & 0.832 & 0.621 & 0.557 & \coloredpercent{25.4} / \coloredpercent{33.1} \\
        \cmidrule(lr){1-6}

        \rowcolor{lightgray}
        \textbf{Average} 
            & \textbf{0.294} & \textbf{0.395} & \textbf{0.299} & \textbf{0.259} & \textbf{\coloredpercent{18.6} / \coloredpercent{29.6}} \\
        \bottomrule
    \end{tabular}
    \end{adjustbox}
    \begin{minipage}{0.7\linewidth}
        {\fontsize{6}{9} \selectfont
            ~ \\[-0.4em]
            Baseline (LC): Baseline results with loop closure enabled. \\[-0.4em]
            Baseline (No LC): Baseline results with loop closure manually deactivated. \\[-0.4em]
            \toa (Seq): Results with 3 sequential base stations (freqseq). \\[-0.4em]
            \toa (Cont): Results with 3 continuous base stations (freq78). \\[-0.4em]
            Improvement (\%) is calculated relative to the baseline without loop closure. \\[-0.4em]
            V203 shows higher error due to tracking losses and relocalization.
        }
    \end{minipage}
\end{table*}

When loop closure is disabled, we observe significant degradation in localization accuracy across most sequences, with the average ATE increasing from 0.294 m to 0.395 m (a 34.35\% degradation). \toa integration effectively mitigates this performance loss, with three sequential base stations improving accuracy by 18.6\% compared to the no-loop-closure baseline. The improvement becomes even more pronounced with continuous \toa measurements from three base stations, achieving a 29.6\% enhancement and reducing the average error from 0.395 m to 0.259 m.

The improvements are particularly striking in challenging sequences where loop closure is typically critical for maintaining baseline performance. For example, in sequence V104, continuous \toa measurements deliver a remarkable 49.3\% improvement in accuracy over the no-loop-closure baseline, while sequence V203 shows a 33.1\% improvement. These significant gains  demonstrate that \toa integration can effectively replace loop closure in preserving trajectory consistency, even without prior knowledge of base station locations. Interestingly, in sequence V203, the baseline without loop closure achieves a slightly better performance (0.832 meters) compared to the baseline with loop closure (0.960 meters). This counterintuitive result may be attributed to suboptimal pose graph optimizations, particularly in this challenging sequence, which involves tracking losses and relocalization events. A similar pattern is observed in dataset V202, where the baseline with loop closure (0.190 meters) marginally outperforms the baseline without loop closure (0.213 meters). In this case, using \toa as a substitute for loop closure provides only minimal improvements or even degradation.

\subsubsection[Comparison with UWB-VO and Other State-of-the-Art]{Comparison with UWB-VO and Other State-of-the-Art Methods}
\begin{table*}[htbp]
    \centering
    \caption[Comparison of SLAM Accuracy on EuRoC Machine Hall Sequences]{Comparison with UWB-VO and State-of-the-Art Methods on EuRoC Machine Hall Dataset}
    \label{tab:mh_comparison}
    \begin{adjustbox}{width=0.6\linewidth}
    \begin{tabular}{l c c c c c c c c}
        \toprule
        \multirow{2}{*}{\textbf{Dataset}} & \multirow{2}{*}{\textbf{UWB-VO}} & \multirow{2}{*}{\textbf{VINS-mono}} & \multirow{2}{*}{\textbf{OKVINS}} & \multirow{2}{*}{\textbf{ROVIO}} & \multicolumn{2}{c}{\textbf{Ours (1BS)}} & \multicolumn{2}{c}{\textbf{Ours (2 Seq BS)}} \\
        \cmidrule(lr){6-7} \cmidrule(lr){8-9}
        & & & & & \textbf{Local} & \textbf{Unscaled} & \textbf{Local} & \textbf{Unscaled} \\
        \midrule

        \multirow{2}{*}{MH01} 
            & \multirow{2}{*}{0.120} & \multirow{2}{*}{0.270} & \multirow{2}{*}{0.160} & \multirow{2}{*}{0.210} 
            & \textbf{0.102} & \multirow{2}{*}{0.116} & 0.103 & \multirow{2}{*}{0.129} \\
            & & & & & \textit{(1.26\%)} &  & \textit{(1.84\%)} & \\
        \cmidrule(lr){1-9}

        \multirow{2}{*}{MH02} 
            & \multirow{2}{*}{0.130} & \multirow{2}{*}{0.150} & \multirow{2}{*}{0.220} & \multirow{2}{*}{0.250} 
            & \textbf{0.119} & \multirow{2}{*}{0.120} & 0.120 & \multirow{2}{*}{0.125} \\
            & & & & & \textit{(0.26\%)} &  & \textit{(0.66\%)} & \\
        \cmidrule(lr){1-9}

        \multirow{2}{*}{MH03} 
            & \multirow{2}{*}{0.280} & \multirow{2}{*}{\textbf{0.130}} & \multirow{2}{*}{0.240} & \multirow{2}{*}{0.250} 
            & 0.208 & \multirow{2}{*}{0.215} & 0.208 & \multirow{2}{*}{0.212} \\
            & & & & & \textit{(1.20\%)} &  & \textit{(0.31\%)} & \\
        \cmidrule(lr){1-9}

        \multirow{2}{*}{MH04} 
            & \multirow{2}{*}{0.360} & \multirow{2}{*}{\textbf{0.250}} & \multirow{2}{*}{0.340} & \multirow{2}{*}{0.490} 
            & 0.289 & \multirow{2}{*}{0.349} & 0.286 & \multirow{2}{*}{0.317} \\
            & & & & & \textit{(2.53\%)} &  & \textit{(1.35\%)} & \\
        \cmidrule(lr){1-9}

        \multirow{2}{*}{MH05} 
            & \multirow{2}{*}{0.300} & \multirow{2}{*}{0.350} & \multirow{2}{*}{0.470} & \multirow{2}{*}{0.520} 
            & 0.300 & \multirow{2}{*}{0.460} & \textbf{0.294} & \multirow{2}{*}{0.375} \\
            & & & & & \textit{(3.93\%)} &  & \textit{(1.82\%)} & \\
        \cmidrule(lr){1-9}

        \multirow{2}{*}{\textbf{Mean}} 
            & \multirow{2}{*}{0.238} & \multirow{2}{*}{0.230} & \multirow{2}{*}{0.286} & \multirow{2}{*}{0.344} 
            & 0.204 & \multirow{2}{*}{0.252} & \textbf{0.202} & \multirow{2}{*}{0.232} \\
            & & & & & \textit{(1.84\%)} &  & \textit{(1.20\%)} & \\
        \bottomrule
    \end{tabular}
    \end{adjustbox}
    \begin{minipage}{0.6\linewidth}
        {\fontsize{6}{9} \selectfont
            ~ \\[-0.4em]    
            1BS: Single base station at [10,10,10] with 5Hz measurements \\[-0.4em]
            2BS: Two sequential base stations active during 20-55s and 60-100s respectively \\[-0.4em]
            Local: Results with scale correction applied \\[-0.4em]
            Unscaled: Results without scale correction \\[-0.4em]
            Bold values indicate the best RMSE performance for each sequence \\[-0.4em]
            Values in italics show scale errors for our methods \\[-0.4em]
            UWB-VO reported scaling errors < 1\% but did not provide detailed results
        }
    \end{minipage}
\end{table*}

Table \ref{tab:mh_comparison} presents a detailed comparison, introducing both Local (scaled) and Unscaled RMSE values. The Local results reflect the accuracy with scale correction applied (Sim(3) alignment), while the Unscaled results show raw SE(3) alignment-based errors, making direct scale comparisons possible.

The UWB-VO approach, proposed by Li et al., is a tightly coupled SLAM framework that integrates monocular visual odometry with UWB \toa constraints to address scale ambiguity and reduce drift. 
The system assumes the presence of a single static UWB anchor and was evaluated using the Machine Hall dataset from the EuRoC MAV benchmark. Notably, their experiments rely on simulated UWB \toa data, generated with a Gaussian distribution at a standard deviation of 5 cm and a frequency of 30Hz.

\begin{figure}[htbp]
    \centering
    \includegraphics[width=1\linewidth]{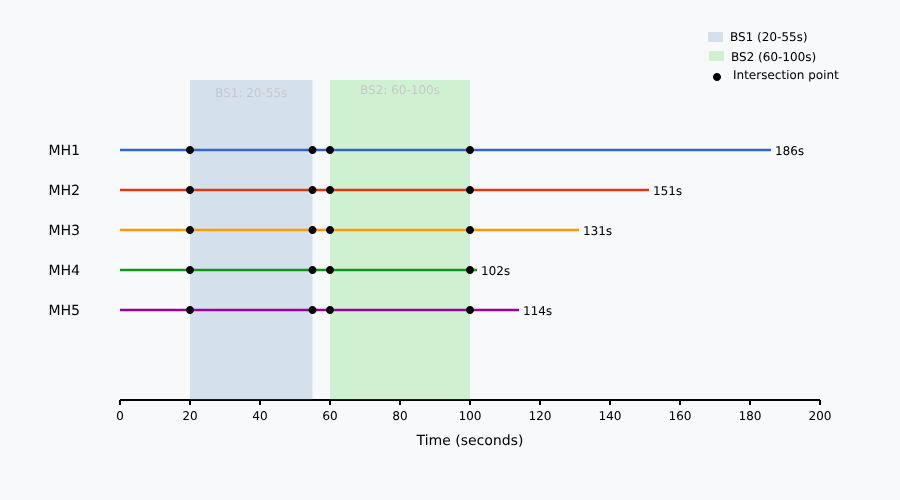}
    \caption[Bastion Station Activity Over Trajectories (Euroc MAV Machine Hall)]{Bastion Station Activity Over Trajectories (Euroc MAV Machine Hall): This plot illustrates the active periods of two bastion stations (BS1: 20-55s, BS2: 60-100s) over the trajectories of Machine Hall (MH01-MH05). Each horizontal bar represents the total duration of a dataset, with colored segments highlighting the intervals during which each bastion station is active}
    \label{fig:mh_timeline}
\end{figure}

Unlike UWB-VO, which relies on a single simulated UWB anchor at 30 Hz, our approach evaluates both single and sequential base station setups (Fig \ref{fig:mh_timeline}) at a lower, more practical 5 Hz measurement rate. The sequential two-base-station approach improves performance in challenging sequences (MH04, MH05).

Despite using a lower \toa measurement frequency, our method achieves a comparable 0.202 m mean RMSE. Even in Unscaled form, the RMSE remains competitive, demonstrating strong inherent scale estimation without external correction. In contrast, UWB-VO reports <1\% scale error but does not provide per-sequence details, making direct comparisons difficult.

These results highlight that our sequential base station method and lower measurement rate still yield robust performance, outperforming several established methods while operating under more relaxed and realistic constraints.

\section{Analysis of Base Station Configurations for SLAM Performance}

In this section, we analyze the impact of the base station position on the SLAM performance.  Geometric Dilution of Precision (GDOP) plays a crucial role in determining localization accuracy through optimal base station positioning. In this section, we analyze the impact of base station configuration on localization performance. GDOP quantifies how the geometric arrangement of base stations affects position estimation accuracy and is calculated as:

\begin{gather}
G = \begin{bmatrix}
\frac{x_1 - x_r}{d_1} & \frac{y_1 - y_r}{d_1} & \frac{z_1 - z_r}{d_1} & 1 \\
\frac{x_2 - x_r}{d_2} & \frac{y_2 - y_r}{d_2} & \frac{z_2 - z_r}{d_2} & 1 \\
\vdots & \vdots & \vdots & \vdots \\
\frac{x_n - x_r}{d_n} & \frac{y_n - y_r}{d_n} & \frac{z_n - z_r}{d_n} & 1
\end{bmatrix}\\
\text{GDOP} = \sqrt{\text{trace}((G^T G)^{-1})}
\end{gather}

\begin{figure*}[h]
    \centering
    \begin{subfigure}[t]{0.19\textwidth}
        \centering
        \includegraphics[width=0.99\textwidth]{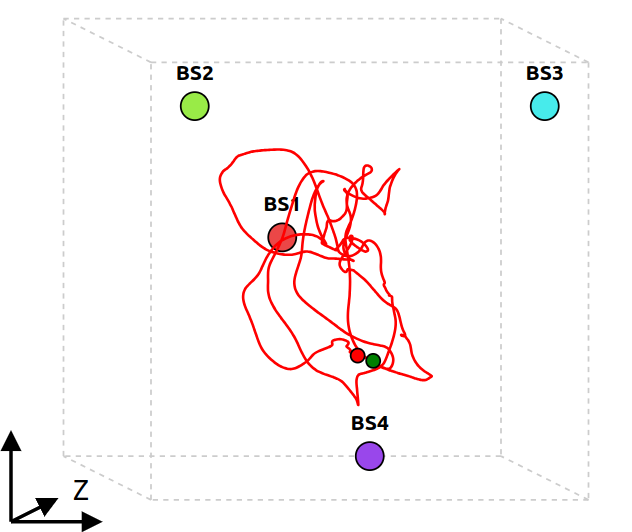}
    \end{subfigure}
    \hfill
    \begin{subfigure}[t]{0.19\textwidth}
        \centering
        \includegraphics[width=0.95\textwidth]{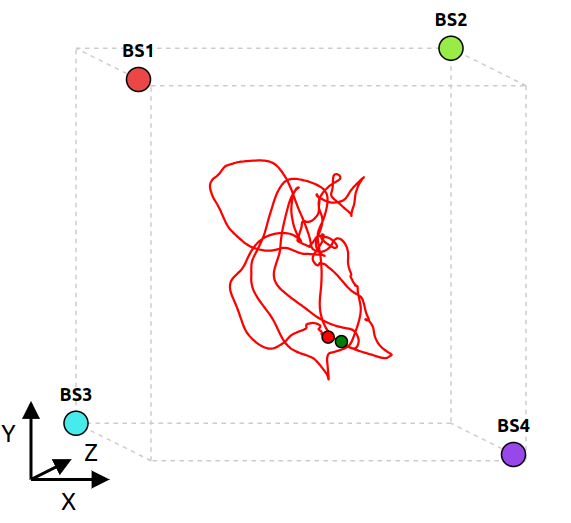}
    \end{subfigure}
    \hfill
    \begin{subfigure}[t]{0.19\textwidth}
        \centering
        \includegraphics[width=\textwidth]{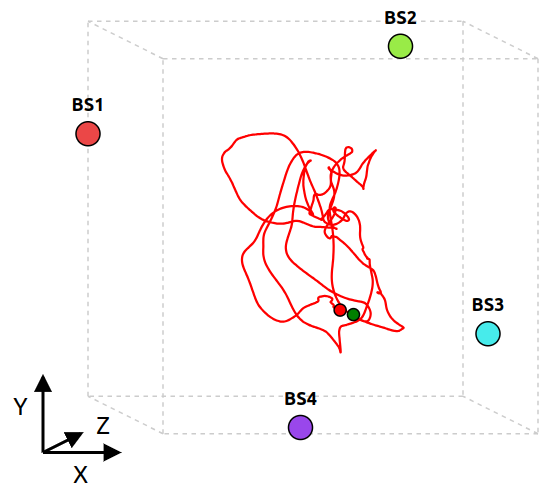}
    \end{subfigure}
    \hfill
    \begin{subfigure}[t]{0.19\textwidth}
        \centering
        \includegraphics[width=\textwidth]{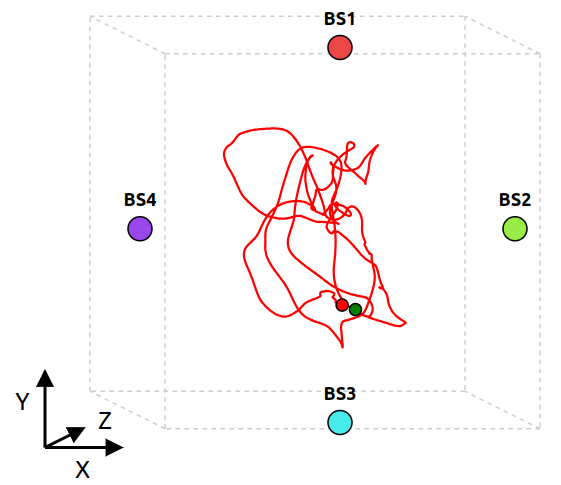}
    \end{subfigure}
    \hfill
    \begin{subfigure}[t]{0.19\textwidth}
        \centering
        \includegraphics[width=\textwidth]{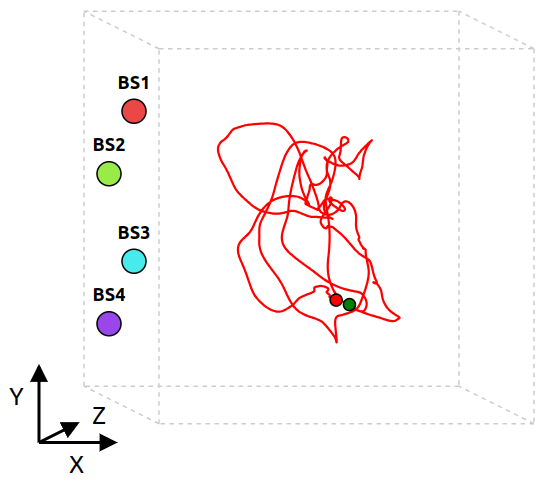}
    \end{subfigure}
    
    \begin{subfigure}[t]{0.19\textwidth}
        \centering
        \includegraphics[width=\textwidth]{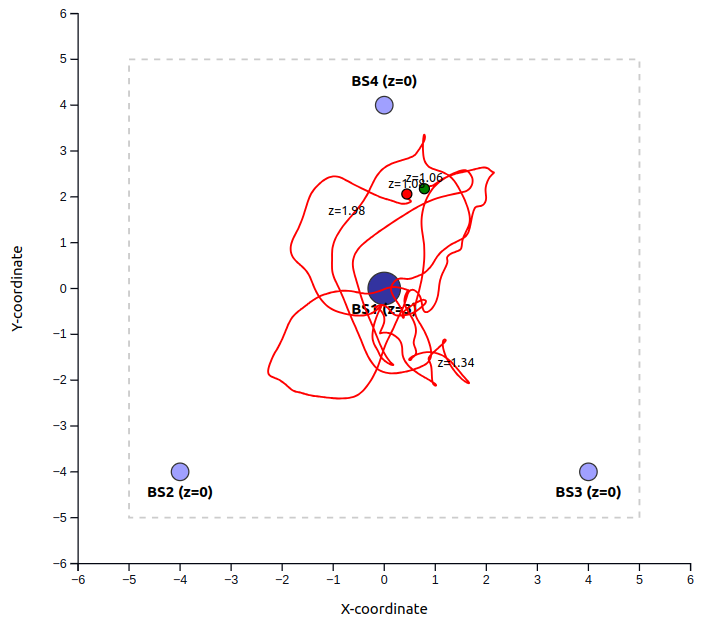}
        \caption{Tetrahedral}
        \label{fig:tetra}
    \end{subfigure}
    \hfill
    \begin{subfigure}[t]{0.19\textwidth}
        \centering
        \includegraphics[width=\textwidth]{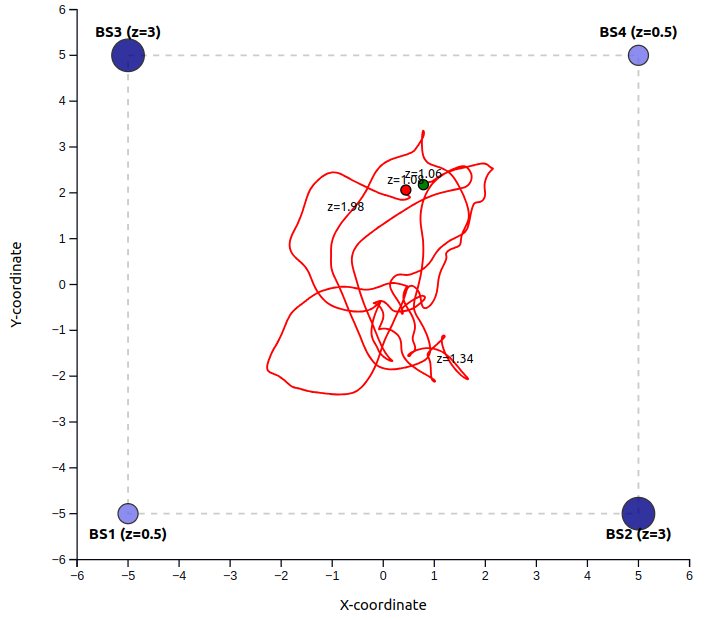}
        \caption{Z-shaped}
        \label{fig:zshaped}
    \end{subfigure}
    \hfill
    \begin{subfigure}[t]{0.19\textwidth}
        \centering
        \includegraphics[width=\textwidth]{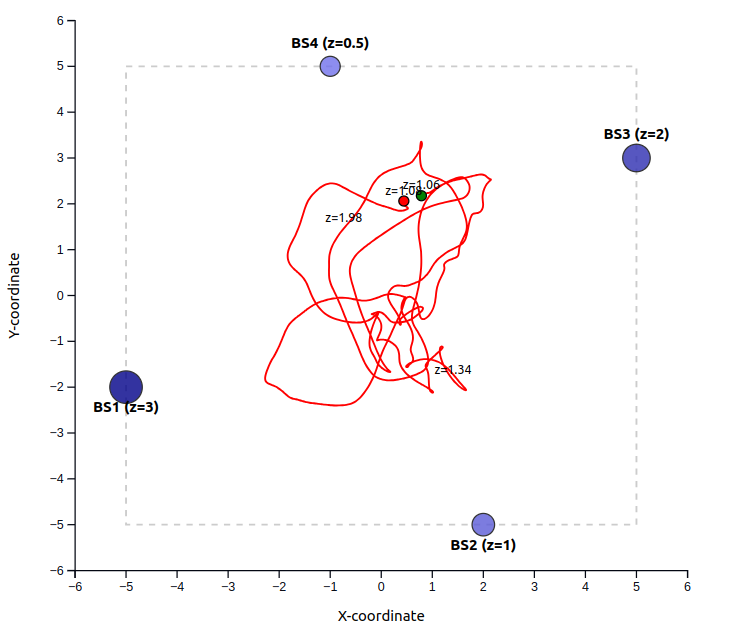}
        \caption{Asymmetric}
        \label{fig:asym}
    \end{subfigure}
    \hfill
    \begin{subfigure}[t]{0.19\textwidth}
        \centering
        \includegraphics[width=\textwidth]{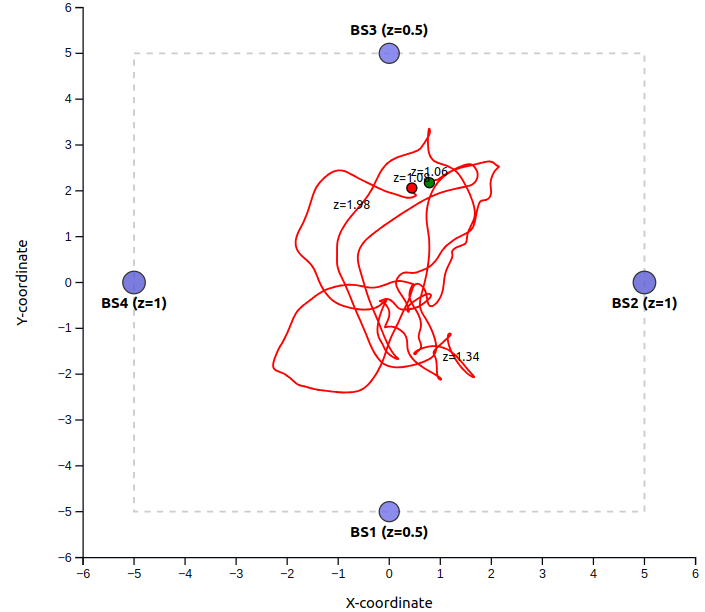}
        \caption{Diamond}
        \label{fig:corner}
    \end{subfigure}
    \hfill
    \begin{subfigure}[t]{0.19\textwidth}
        \centering
        \includegraphics[width=\textwidth]{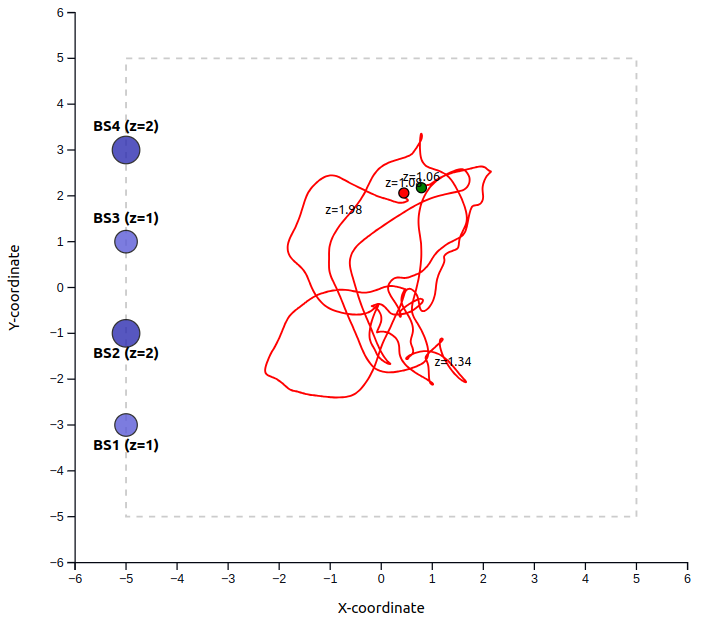}
        \caption{Clustered}
        \label{fig:clustered}
    \end{subfigure}
    
    \caption[Five configurations of base stations (BS1-BS4)]{Five configurations of base stations (BS1-BS4) with trajectory data from the EuRoC MAV V101 dataset. Top row: 3D representations; Bottom row: corresponding 2D X-Y projections with z-coordinates. Configurations are ordered by performance (best to worst): Tetrahedral (ATE: 0.111 m), Z-shaped (0.132 m), Asymmetric (0.141 m), Corner (0.161 m), and Clustered (0.182 m).}
    \label{fig:all_configurations}
\end{figure*}

Lower GDOP values, achieved when a mobile device is surrounded by well-distributed base stations, correspond to higher positioning accuracy, while clustered configurations result in higher GDOP and degraded performance. Our analysis examines five representative base station arrangements (Tetrahedral, Asymmetric, Diamond, Z-Shape, and Clustered) to demonstrate how geometric configuration directly impacts localization precision, particularly in applications requiring high accuracy such as drone navigation. For a deeper exploration of GDOP and its optimization, readers are referred to \cite{sharp2009gdop} and \cite{5467147}.


we use five distinct base station configurations (Fig.\ref{fig:all_configurations}) with exact coordinates detailed in Table \ref{tab:bs_coordinates}:

\textbf{Tetrahedral}: 3D distributed (ceiling center + ground triangle)

\textbf{Z-Shape}: Diagonally alternating heights

\textbf{Asymmetric}: Realistic irregular placement

\textbf{Diamond}: Distributed evenly at two different heights

\textbf{Clustered}: Single-wall co-planar arrangement

\begin{table}[h]
\centering
\caption{Exact 3D coordinates (in meters) of base stations for all evaluated configurations.}
\label{tab:bs_coordinates}
\begin{adjustbox}{width=1\linewidth}
\begin{tabular}{lccccc}
\toprule
\textbf{Base Station} & \textbf{Tetrahedral} & \textbf{Diamond} & \textbf{Z-Shape} & \textbf{Asymmetric} & \textbf{Clustered} \\
\midrule
BS1 & (0, 0, 3) & (0, -5, 0.5) & (-5, -5, 0.5) & (-5, -2, 3) & (-5, -3, 1) \\
\rowcolor{lightgray}
BS2 & (-4, -4, 0) & (5, 0, 1) & (5, -5, 3) & (2, -5, 1) & (-5, -1, 2) \\
BS3 & (4, -4, 0) & (0, 5, 0.5) & (-5, 5, 3) & (5, 3, 2) & (-5, 1, 1) \\
\rowcolor{lightgray}
BS4 & (0, 4, 0) & (-5, 0, 1) & (5, 5, 0.5) & (-1, 5, 0.5) & (-5, 3, 2) \\
\bottomrule
\end{tabular}
\end{adjustbox}
\end{table}

These configurations were evaluated across both EuRoC MAV and Aerolab Dataset for Stereo and RGB-D respectively. 

\subsection{Results on EuRoC MAV Dataset}

Table~\ref{tab:euroc_rmse_configs} presents the average ATE values for both local and global accuracy across all configurations and EuRoC MAV dataset sequences.

\begin{table}[h]
\centering
\caption[ATE for different base station configurations with 5G \toa integrated stereo across EuRoC MAV datasets]{ATE for different base station configurations across EuRoC MAV datasets (meters). Each cell contains local accuracy / global accuracy values.}
\label{tab:euroc_rmse_configs}
\begin{adjustbox}{width=\linewidth}
\begin{tabular}{lcccccc}
\toprule
Configuration & V101 & V102 & V103 & V201 & V202 & Average \\
\midrule
Tetrahedral & 0.078/0.083 & 0.124/0.135 & 0.140/0.158 & 0.124/0.158 & 0.184/0.189 & 0.130/0.145 \\
\rowcolor{lightgray}
Z-Shape & 0.079/0.082 & 0.108/0.129 & 0.177/0.203 & 0.087/0.135 & 0.188/0.199 & 0.128/0.150 \\
Asymmetric & 0.079/0.083 & 0.105/0.118 & 0.186/0.239 & 0.131/0.207 & 0.183/0.203 & 0.137/0.170 \\
\rowcolor{lightgray}
Diamond & 0.078/0.083 & 0.186/0.286 & 0.148/0.162 & 0.084/0.118 & 0.177/0.188 & 0.135/0.167 \\
Clustered & 0.084/0.095 & 0.137/0.155 & 0.223/0.628 & 0.096/0.245 & 0.189/0.204 & 0.146/0.265 \\
\bottomrule
\end{tabular}
\end{adjustbox}
\end{table}

The results demonstrate the impact of the base station geometry on both local and global SLAM accuracy, with the Z-Shape configuration showing the best average local accuracy (0.128 m) and the Tetrahedral configuration achieving the best average global accuracy (0.145 m). The Clustered configuration showed the poorest performance in both metrics, with a particularly significant degradation in global accuracy (0.265 m, 82.8\% worse than the Tetrahedral configuration).

\subsection{Results on AeroLab Dataset}
 The same evaluation of different base station configurations was conducted using the AeroLab dataset. Table~\ref{tab:aerolab_rmse_configs} presents the average ATE values.

\begin{table}[h]
\centering
\caption[ATE for different base station configurations across AeroLab datasets]{ATE for different base station configurations with 5G \toa integrated RGB-D across AeroLab datasets (meters). Each cell contains local accuracy / global accuracy values.}
\label{tab:aerolab_rmse_configs}
\begin{adjustbox}{width=\linewidth}
\begin{tabular}{lcccccc}
\toprule
Configuration & AL01 & AL02 & AL03 & AL04 & AL05 & Average \\
\midrule
Tetrahedral & 0.063/0.092 & 0.110/0.115 & 0.143/0.150 & 0.092/0.095 & 0.165/0.166 & 0.115/0.124 \\
\rowcolor{lightgray}
Z-Shape & 0.062/0.095 & 0.107/0.125 & 0.146/0.155 & 0.095/0.105 & 0.157/0.165 & 0.113/0.129 \\
\rowcolor{lightgray}
Asymmetric & 0.068/0.112 & 0.109/0.142 & 0.155/0.160 & 0.099/0.103 & 0.172/0.176 & 0.121/0.138 \\
\rowcolor{lightgray}
Diamond & 0.062/0.097 & 0.110/0.138 & 0.143/0.152 & 0.097/0.117 & 0.165/0.172 & 0.115/0.135 \\
Clustered & 0.069/0.131 & 0.113/0.144 & 0.148/0.159 & 0.091/0.096 & 0.176/0.184 & 0.119/0.143 \\
\bottomrule
\end{tabular}
\end{adjustbox}
\end{table}

The AeroLab results show almost the same pattern as the EuRoC dataset. The Z-Shape configuration achieved the best average local accuracy (0.113 m), closely followed by the Tetrahedral and Diamond configurations (both 0.115 m). For global accuracy, the Tetrahedral configuration maintained its advantage (0.124 m), with the Z-Shape configuration a close second (0.129 m). The Clustered configuration again showed the poorest performance overall, particularly in global accuracy (0.143 m).

\subsection{Cross-Dataset Analysis}

Comparing the results across both datasets reveals several consistent patterns and some interesting differences, as shown in Table~\ref{tab:combined_rmse}.

\begin{table}[h]
\centering
\caption{Combined average RMSE across both datasets (meters)}
\label{tab:combined_rmse}
\begin{adjustbox}{width=1\linewidth}
\begin{tabular}{lcc c}
\toprule
Configuration & EuRoC Local/Global & AeroLab Local/Global & Combined Local/Global \\
\midrule
Tetrahedral & 0.130/0.145 & 0.115/0.124 & 0.123/0.135 \\
\rowcolor{lightgray}
Z-Shape & 0.128/0.150 & 0.113/0.129 & 0.121/0.140 \\
Diamond & 0.135/0.167 & 0.115/0.135 & 0.125/0.151 \\
\rowcolor{lightgray}
Asymmetric & 0.137/0.170 & 0.121/0.138 & 0.129/0.154 \\
Clustered & 0.146/0.265 & 0.119/0.143 & 0.133/0.204 \\
\bottomrule
\end{tabular}
\end{adjustbox}
\end{table}

The combined results highlight several key insights:

\begin{itemize}
    \item The Z-Shape configuration shows the best overall local accuracy (0.121m), with the Tetrahedral configuration a close second (0.123m).
    \item The Tetrahedral configuration delivers the best overall global accuracy (0.135m), outperforming other configurations by at least 3.7\%.
    \item The Clustered configuration performs consistently worse across both datasets, with significantly degraded global accuracy (0.204m, 51.1\% worse than Tetrahedral).
    \item The performance gap between configurations is more pronounced in the EuRoC dataset than in the AeroLab dataset, suggesting that environment complexity and motion dynamics play important roles in determining the impact of base station geometry.
\end{itemize}

\subsection{Global-to-Local Accuracy Ratio}

The ratio between global and local accuracy provides insight into how well each configuration maintains global consistency. Table~\ref{tab:global_local_ratio} presents these ratios for both datasets.

\begin{table}[h]
\centering
\caption{Global-to-local accuracy ratio for different base station configurations}
\label{tab:global_local_ratio}
\begin{tabular}{lccc}
\toprule
Configuration & EuRoC Ratio & AeroLab Ratio & Combined Ratio \\
\midrule
Tetrahedral & 1.12 & 1.08 & 1.10 \\
\rowcolor{lightgray}
Z-Shape & 1.17 & 1.14 & 1.16 \\
Diamond & 1.24 & 1.17 & 1.21 \\
\rowcolor{lightgray}
Asymmetric & 1.24 & 1.14 & 1.19 \\
Clustered & 1.82 & 1.20 & 1.53 \\
\bottomrule
\end{tabular}
\end{table}

The Tetrahedral configuration shows the lowest ratios across both datasets (1.10 combined), indicating that its local trajectory accuracy translates well to global positioning. The Clustered configuration has the highest ratio (1.53 combined), demonstrating significant degradation when moving from local to global accuracy assessment. This degradation is particularly severe in the EuRoC dataset (ratio of 1.82), highlighting the vulnerability of the Clustered configuration to challenging conditions.

\subsection{Relationship Between GDOP and SLAM Performance}

\begin{figure}[h]
    \centering
    \includegraphics[width=\linewidth]{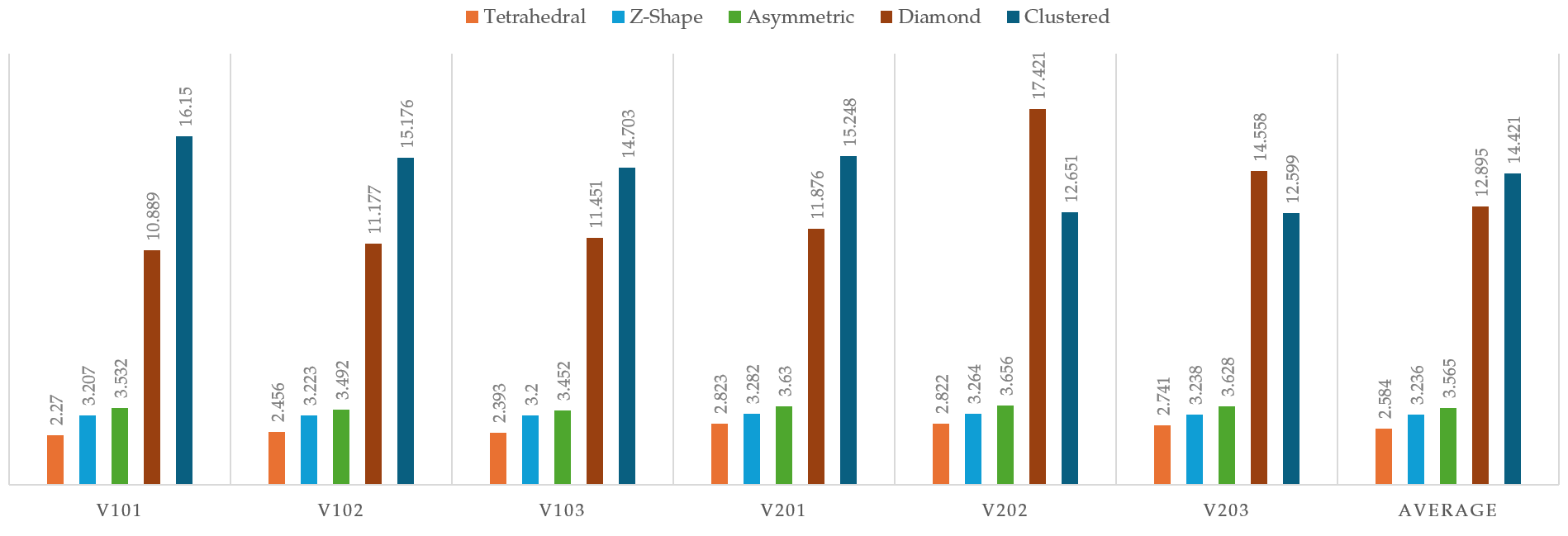}
    \caption[Comparison of average Geometric Dilution of Precision (GDOP)]{Comparison of average Geometric Dilution of Precision (GDOP) values across different base station configurations for all EuRoC MAV dataset sequences. Lower GDOP values indicate better geometric configuration for localization.}
    \label{fig:gdop_configurations}
\end{figure}

\begin{figure}[h!]
    \centering
    \includegraphics[width=\linewidth]{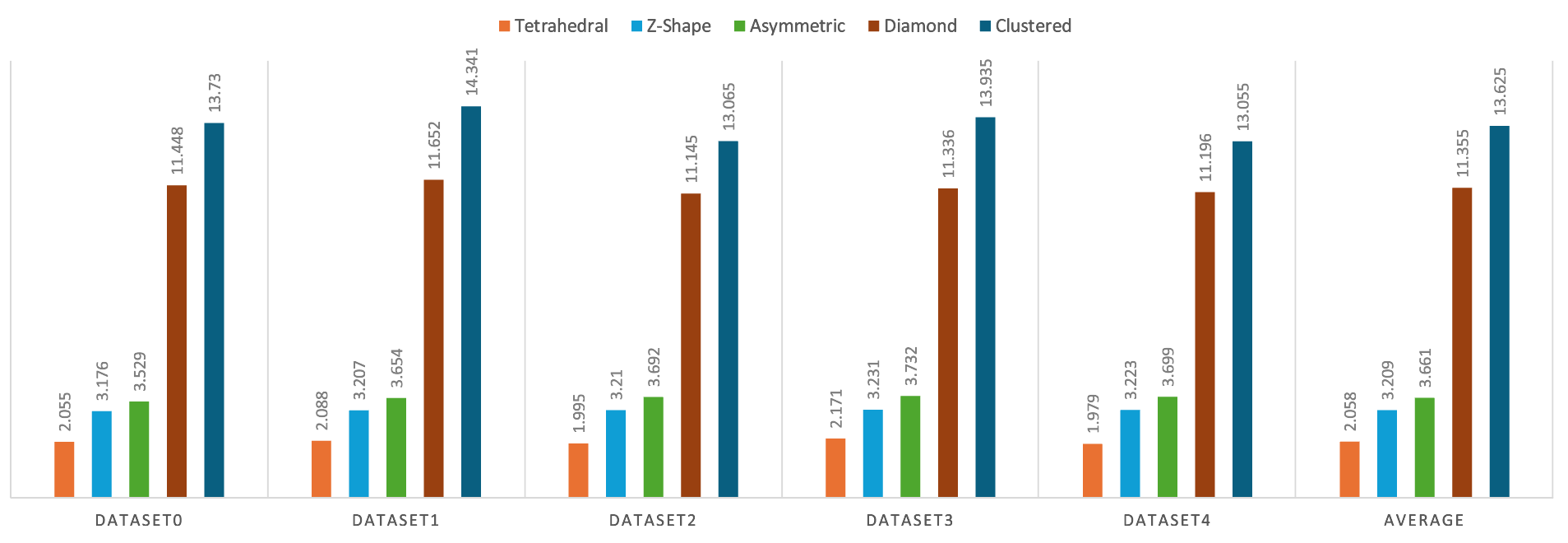}
    \caption[Comparison of average Geometric Dilution of Precision (GDOP) for AeroLab]{Comparison of average Geometric Dilution of Precision (GDOP) values across different base station configurations for all AeroLab dataset sequences. Lower GDOP values indicate better geometric configuration for localization.}
    \label{fig:gdop_configurations_aerolab}
\end{figure}

The bar chart in Figure~\ref{fig:gdop_configurations} and \ref{fig:gdop_configurations_aerolab} presents the average GDOP values for each base station configuration across all EuRoC MAV dataset sequences and Aerolab respectively, providing direct empirical evidence of how geometric arrangement affects positioning quality. A correlation between GDOP values and SLAM performance across both datasets is evident especially for global accuracy. The Tetrahedral configuration's superior global accuracy directly corresponds to its lowest average GDOP in both datasets. However, the fusion with visual data helps mitigate some of the geometric limitations, especially for local trajectory accuracy, resulting in smaller percentage differences between configurations compared to pure localization.


\section{System Limitations and Future Enhancements}
\label{sec:limitation}

A fundamental consideration of this work, though inherent to \toa-based positioning rather than our framework's design, is the assumption of fixed and precisely known base station positions. Firstly, this assumption is crucial for enabling global SLAM capabilities, as these stations serve as critical global references. Global SLAM fundamentally relies on at least some fixed and known reference points to establish absolute positioning. While it is theoretically possible to estimate base station positions dynamically, such approaches typically confine SLAM to local frames, thereby limiting its global applicability. Even if we consider only local SLAM and neglect global positioning requirements, another challenge emerges: Unlike traditional SLAM landmarks, which provide rich 3D or SE3 (6-DoF) measurements and can self-correct during optimization, \toa measurements are scalar and offer only radial distance constraints. This fundamental difference creates observability challenges, as each \toa reading constrains the base station position to a sphere around the UAV. While multiple measurements along a trajectory create intersecting spheres that could theoretically pinpoint base station positions, in practice \toa measurement accuracy (typically 20-30 cm) expands each sphere into a thick shell. This measurement uncertainty, rather than geometric ambiguity, is what makes reliable convergence challenging without accurate initial position estimates for both the drone's pose and base station positions. This significantly complicates system initialization and contrasts with the current framework, which assumes that the drone's initial pose is unknown.


In practical deployment scenarios, this limitation is mitigated by the nature of indoor 5G infrastructure, which typically involves fixed installations and one-time deployment. Calibrating base station positions during setup ensures consistent global localization while significantly reducing system complexity. While some applications may involve dynamic or unknown base station positions, maintaining a subset of known positions as global anchors remains essential for establishing a consistent global reference frame. This trade-off between practicality and flexibility enables robust global SLAM while addressing key challenges associated with scalar \toa measurements.

Beyond the base station positioning challenges, several other limitations affect the system's real-world applicability. The current \toa simulation framework assumes ideal Line-of-Sight (LoS) conditions, while  Non-Line-of-Sight (NLoS) conditions, caused by obstacles and reflections, can significantly degrade \toa measurement accuracy and impact SLAM performance. This limitation is compounded by the use of static information matrices for both \toa and visual-inertial measurements, which fail to capture the dynamic nature of measurement uncertainties in real environments. Moreover, while simulation provides a controlled testing environment, it may not fully capture the complexities of real-world signal propagation, hardware noise, and environmental dynamics. These limitations highlight the need for future validation with experimental \toa measurements and the development of more sophisticated uncertainty models that can adapt to changing environmental conditions.
\section{Conclusion}
\label{sec:conclusion}
This paper presented a novel approach for integrating 5G Time of Arrival (\toa) measurements into ORB-SLAM3, enabling globally consistent localization and mapping for indoor drone navigation. By extending ORB-SLAM3's optimization pipeline to jointly process \toa data alongside visual and inertial measurements, we successfully transformed inherently local SLAM estimates into globally referenced trajectories while effectively resolving scale ambiguity in monocular configurations.
Our comprehensive evaluation across multiple datasets and SLAM configurations demonstrated that \toa integration maintains high local accuracy while enabling consistent global positioning - a capability absent in traditional SLAM systems. For monocular systems specifically, the integration of \toa measurements successfully addressed the fundamental challenge of scale ambiguity without requiring additional sensors or manual calibration.

The investigation into unknown base station positions revealed that our approach remains effective even with limited knowledge of anchor positions, showcasing the system's adaptability to real-world deployment constraints. Furthermore, we demonstrated that \toa measurements can serve as a viable alternative to traditional loop closure mechanisms, maintaining trajectory consistency in scenarios where loop detection might be unreliable or impossible.
Our analysis of different geometric arrangements of base stations provided valuable insights into optimal deployment strategies, revealing performance differences between configurations. This knowledge directly informs practical implementation by guiding base station placement decisions.

Comparative analysis with state-of-the-art methods confirmed the robustness of our approach even under challenging operational constraints such as lower measurement frequencies and sequential base station operation. The experimental results collectively validate that 5G \toa integration offers substantial benefits for global SLAM applications, particularly in challenging indoor environments where accurate positioning is critical.
Future work could explore real-world implementation with physical 5G infrastructure, investigate dynamic base station calibration techniques, and extend the approach to multi-agent collaborative mapping scenarios.

\section*{Acknowledgment}

The authors would like to thank Hamed Habibi, Mohan Dasari, and Pedro Soares for their invaluable assistance with the experimental setup, flight operations, and data collection. Their support was pivotal to the success of this study.


\bibliographystyle{IEEEtran}
\bibliography{refs_all}

\begin{IEEEbiography}[{\includegraphics[width=1in,height=1.25in,clip,keepaspectratio]{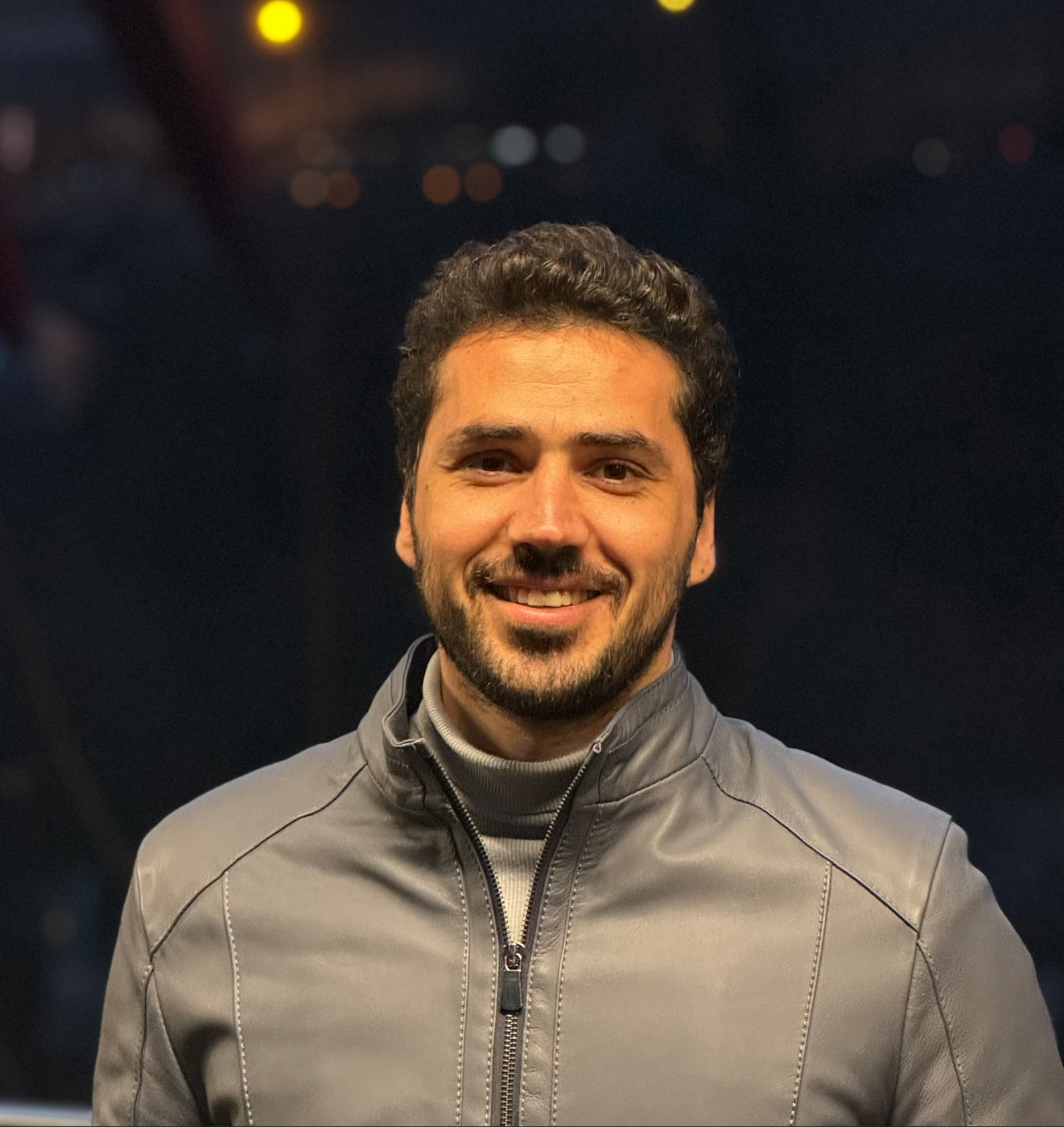}}]{Meisam Kabiri} 
received his M.Sc. degree in Electrical Engineering from Amirkabir University of Technology (Tehran Polytechnic), Tehran, Iran, in 2017. He is currently a PhD researcher at the University of Luxembourg, affiliated with the Interdisciplinary Centre for Security, Reliability and Trust (SnT). His research focuses on mobile robotics, particularly in the areas of Simultaneous Localization and Mapping (SLAM) and distributed networked control systems. Meisam is a member of the Automation \& Robotics Research Group (ARG) at SnT, under the supervision of Prof. Holger Voos.
\end{IEEEbiography}
\begin{IEEEbiography}[{\includegraphics[width=1in,height=1.25in,clip,keepaspectratio]{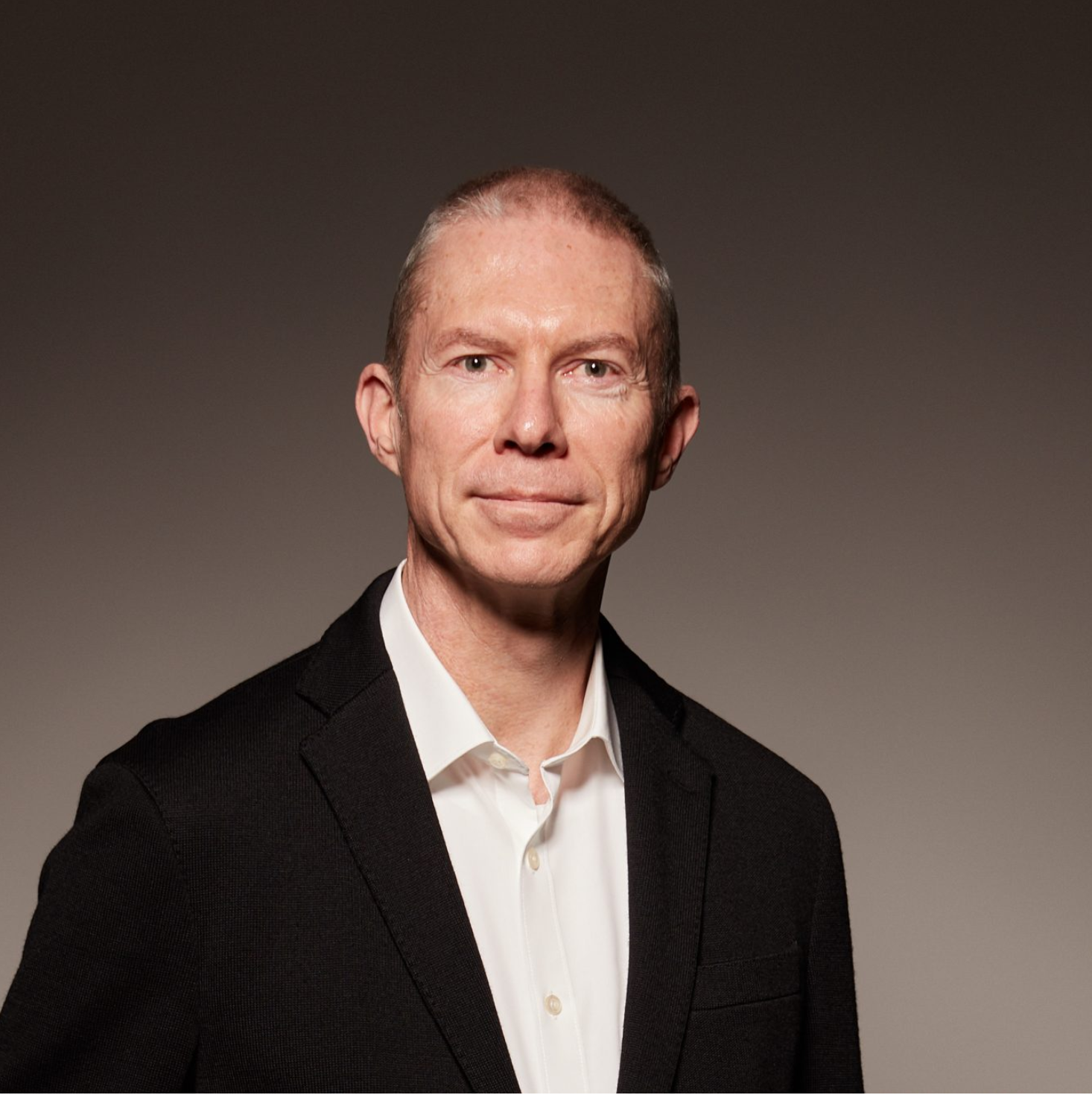}}]{Holger Voos}
studied Electrical Engineering at the Saarland University and received the Doctoral Degree in Automatic Control from the Technical University of Kaiserslautern, Germany, in 2002. From 2000 to 2004, he was with Bodenseewerk Gerätetechnik GmbH, Germany, where he worked as a systems engineer and project manager in R\&D in aerospace and robotics. From 2004 to 2010, he was a Professor at the University of Applied Sciences Ravensburg-Weingarten, Germany, and the head of the Mobile Robotics Lab there. Since 2010, he is a Full Professor at the University of Luxembourg in the Interdisciplinary Centre for Security, Reliability and Trust (SnT), and the head of the SnT Automation and Robotics Research Group. His research interests are in the area of perception, situational awareness as well as motion planning and control for autonomous vehicles and robots as well as distributed and networked control and automation. Areas of application are robotics, space systems, Industry 4.0 and energy and water networks. He is author or co-author of more than 300 publications, comprising books, book chapters and journal and conference papers.
\end{IEEEbiography}

\EOD

\end{document}